\documentclass[12pt,journal,final]{IEEEtran}  
\usepackage[T1]{fontenc} 
\usepackage{graphicx}
\usepackage{caption}
\usepackage{subcaption}
\usepackage{amsmath}
\usepackage{xcolor}
\usepackage{multirow}
\usepackage{adjustbox}
\usepackage{multicol}
\usepackage[ruled,vlined]{algorithm2e}
\usepackage[cmintegrals]{newtxmath}
\usepackage{booktabs}
\usepackage{bm} 
\usepackage{footnote}
\makesavenoteenv{tabular}
\makesavenoteenv{table}

\author{\IEEEauthorblockN{Yongliang Chen}
\IEEEauthorblockA{
College of Optoelectronic Engineering\\
Chongqing University, China\\
Email: chen\_yong\_liang@cqu.edu.cn}
}

\title{A Labeling-Free Approach to Supervising Deep Neural Networks 
for Retinal Blood Vessel Segmentation }

\begin{document}
\maketitle
\begin{abstract}
Segmenting blood vessels in fundus imaging plays an important
role in medical diagnosis. Many algorithms have been proposed.
While deep Neural Networks 
have been attracting 
enormous attention from computer vision community recent years and
several novel works have been done in terms of its application
in retinal blood vessel segmentation, most of them are based on 
supervised learning which requires amount of labeled data, which 
is both scarce and expensive to obtain.
We leverage the power of Deep Convolutional 
Neural Networks (DCNN) 
in feature learning,
 in this work, to achieve this ultimate goal.
 The highly efficient feature learning of DCNN
inspires our novel approach that 
trains the networks with 
automatically-generated samples to achieve
desirable performance on real-world fundus images.
For this, we design a set of rules
 abstracted from the domain-specific prior knowledge to generate these samples.
 We argue that, with the high efficiency of DCNN in feature learning,
 one can achieve this goal by constructing the training dataset with  
 prior knowledge, no manual labeling is needed.
This approach allows us to take advantages of 
supervised learning without labeling.
We also build a naive DCNN model to test it.
The results on standard benchmarks of fundus imaging
show it is competitive to the state-of-the-art
methods which implies a potential way to leverage the power 
of DCNN in feature learning.


 \end{abstract}
\section{INTRODUCTION}
Retinal is a key component for  
our visual perception. 
It transforms incoming light to neural signal 
for further processing.
Morphological features of retinal vessel 
can be used for various purposes such as
monitoring the disease progression, treatment, 
and evaluation of various cardiovascular and
ophthalmologic diseases
\cite{melinscak2016retinal}.

While the pattern of retinal vessels delivers significant 
information for diagnosis \cite{abramoff2010retinal},
comprising arteries and lots of veins 
make manual segmentation both tedious and time-consuming
\cite{Fu2016}.
Aimed to develop algorithms for vessel segmentation,
it has been a main focus over years to exploit computer's 
strong ability in computing to
facilitate this work \cite{Matsui1973}.
But complexities existing in retinal images such as
nonuniform illumination,
lower contrast \cite{liskowski2016segmenting},
abrupt variation in branching patterns
\cite{2016arXiv161102064D} makes it a nontrivial task
to accomplish \cite{liskowski2016segmenting}.

Although there has been a number of previous works related to 
this topic \cite{Fraz2012407}. The Artificial Neural Networks, 
namely, Deep Learning \cite{Lecun2015Deep}, 
has been attracting more and more attention
 following the novel works by 
\cite{hinton2006reducing}.
Inspired by the organization of the animal visual
 cortex \cite{hubel1968receptive}, 
Convolutional 
Neural Networks (CNN)
 has made a number of achievements in 
the field of computer visual \cite{Lecun2015Deep}. It is composed of multiple
processing layers, typically, filter bank layer, nonlinearity layer,
feature pooling layer \cite{lecun2010convolutional},
 to learn inherent characters of data 
with multiple levels of abstraction.
It overcomes the previously existed 
drawbacks in processing natural data in their raw form and
achieves competitive performance in pattern recognition \cite{Lecun2015Deep}.
 
In the scope of CNN, supervised learning strategy is commonly invoked
 in training stage \cite{Lecun2015Deep}, in which the learning system 
takes samples as input, and then
 measures the error between current output and its corresponding 
goal, known as label. Then gradient of the 
parameters (weight and bias) in the top layer which is
nearest to the output is 
calculated. By exploiting back-propagation \cite{rumelhart1986learning},
configuration of each layer below the top will be rectified.
CNN with deep structure (Deep Convolutional Neural Networks,
DCNN) have attracted much more attention 
since impressive results have been presented by \cite{krizhevsky2012imagenet}.
It comes with much more layers \cite{he2015deep}, 
more powerful processor (graphics processing unit),
 and larger dataset \cite{deng2009imagenet} and is to achieve 
 unprecedented performance (\cite{krizhevsky2012imagenet} et al.).

As mentioned above, labels are required in supervised learning 
for each input. As deep learning takes more advantages on large
datasets \cite{Bengio2007Scaling}, more labels are desired.
However it is difficult to obtain massive samples together with the
corresponding labels in terms of retinal vessel segmentation;
further more, lack of labeled data becomes a major problem 
hampering us from utilizing its power for specific domain.

Our contributions are summarized as followings:
\begin{itemize}
\item we propose a novel approach to address the lack of labeled data by 
automatically generating samples and labels  from domain-specific knowledge;
\item we also provide a deep neural networks model to evaluate our approach.
\end{itemize}


\section{RELATED WORKS}
In previous works, \cite{marin2011a} prepares the 
feature combining gray-level and statistical moment,
 then neural networks have been applied for 
 pixel classification. This method falls in conventional
 supervised learning.
 
As for deep neural networks, \cite{shelhamer2016fully} proposes a fully 
 convolutional neural networks model for
 image segmentation, providing a framework for pixel-level
 segmentation using DCNN.
 
 Inspired by this, recently, several methods have been proposed for retinal vessel segmentation
 in terms of deep learning.
  \cite{2016arXiv161102064D} formulates the segmentation task as a 
  multi-label inference problem and exploits the dependencies 
  among neighboring pixels to improve the performance.
  \cite{Fu2016} utilizes a conditional random field to 
  model the long-range interaction between pixels.
  \cite{melinscak2016retinal} uses a Graphic Process Unit (GPU)
  implementation of deep neural networks to demonstrate its 
  high effectiveness in segmentation.
  \cite{liskowski2016segmenting} conducts a comprehensive
  study in retinal vessel 
  segmentation using deep neural networks under supervised 
  learning. In their work, the system is fed with 
  small patches cropped from a large one for training, before which 
  the images have been preprocessed. A range of network 
  architectures have also been evaluated in that work; among them, 
  a fully connected layer is widely applied before the 
  final output layer.
  
  All these above are marked as supervised learning approaches,
   and, as mentioned in \cite{Fraz2012407}, 
   they are likely to achieve better
  performance than their unsupervised counterparts.
  We also notice that manually labeled samples are 
  required for all of them.
  
  Still, there is a semi-supervised scheme 
  in which multiple  stacked denoised 
   autoencoders (SDAE) have been trained
  to learn dictionary of visual kernels 
  for segmentation \cite{Lahiri2016ensemble},
  in which an additional layer is need to fuse each one's output;
 However, as a paradigm of semi-supervised learning, 
  labeled data is required in the final stage (fine-tuning).

	\section{PROPOSED METHOD}
\begin{figure}
\centering
\begin{subfigure}[b]{0.23\textwidth}
\includegraphics[scale=0.22]{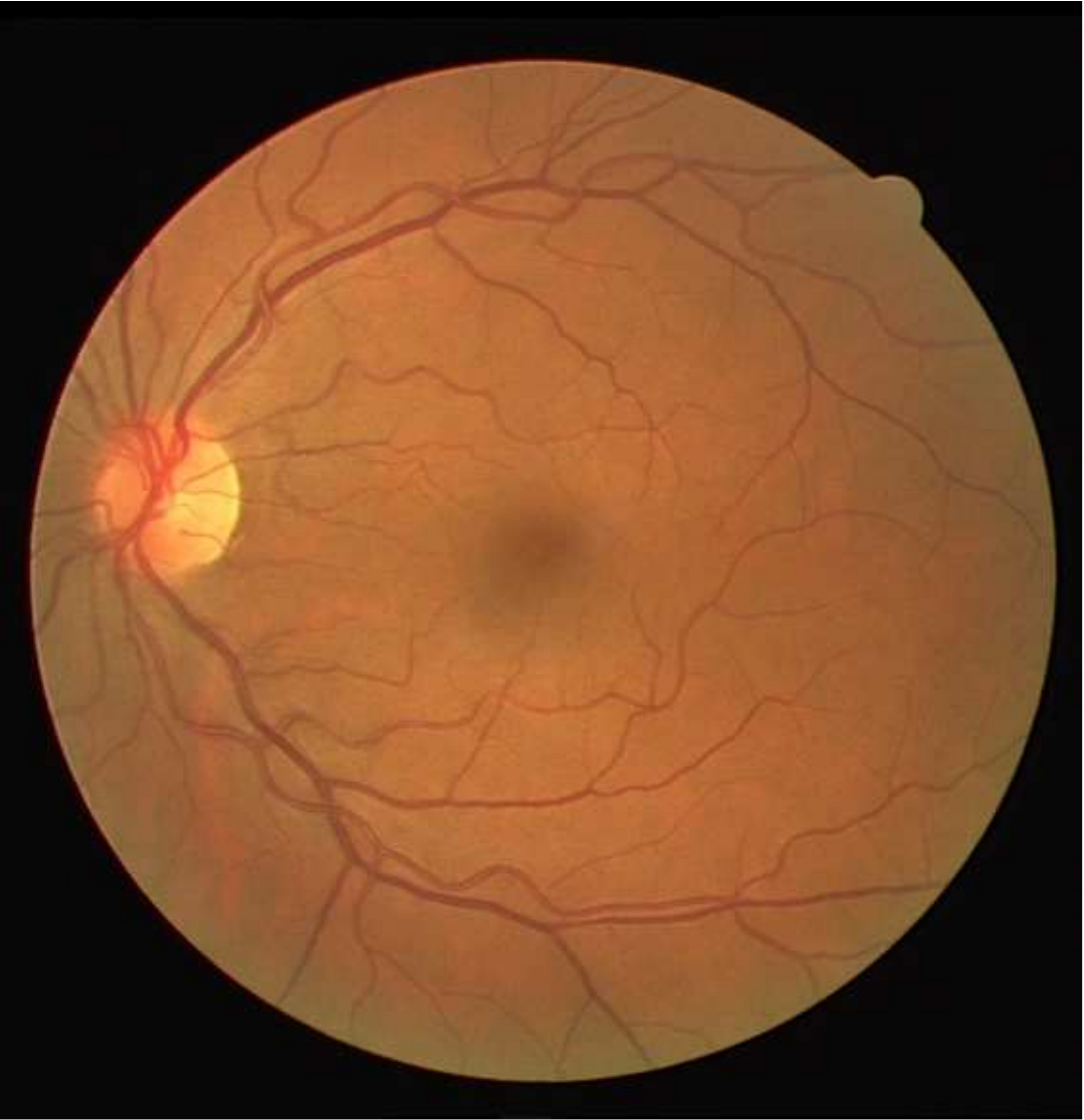}
\end{subfigure}
~
\begin{subfigure}[b]{0.23\textwidth}
\includegraphics[scale=0.22]
{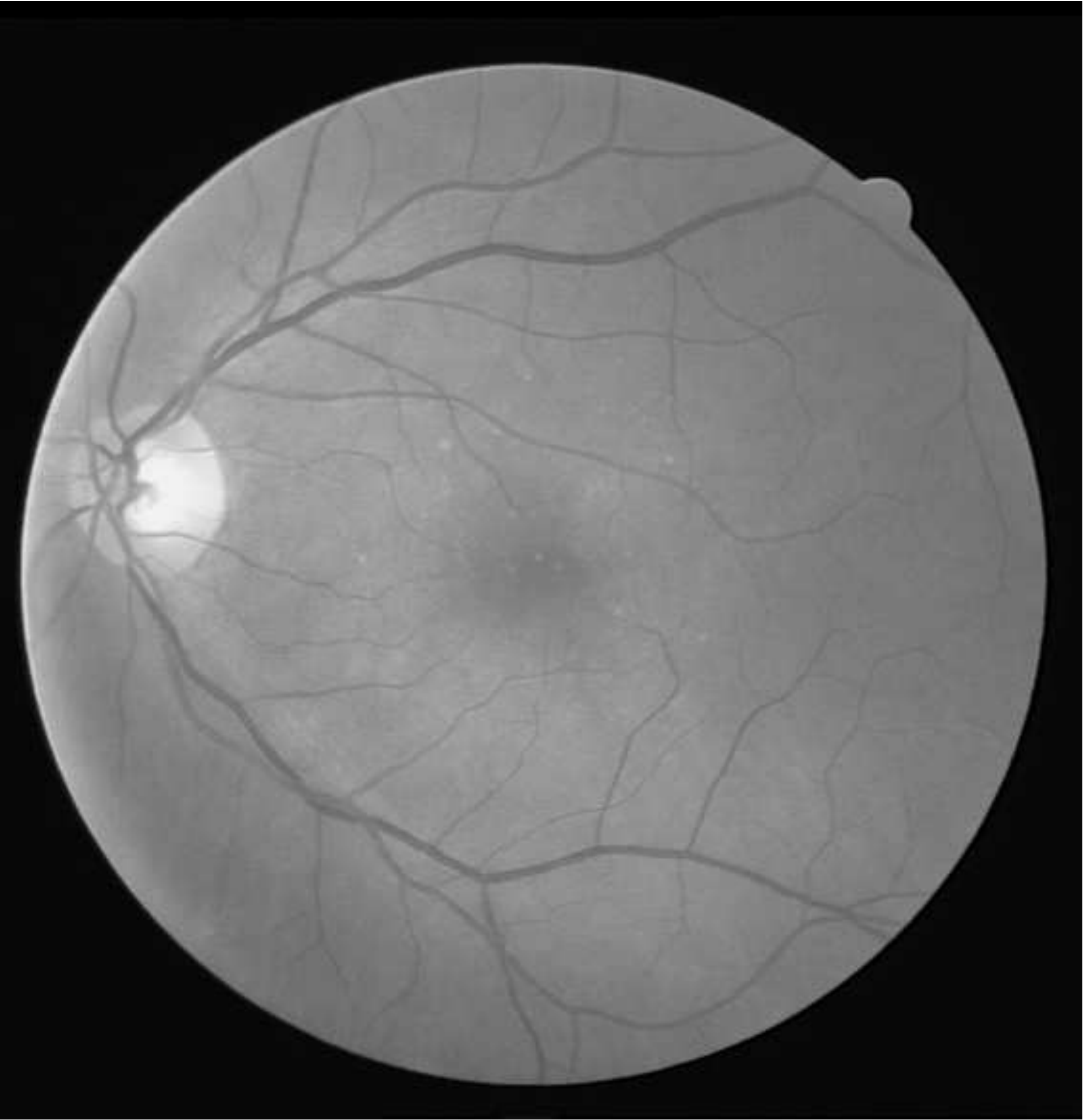}
\end{subfigure}
\caption{A comparison between color image (left) and 
its grayscale version (right).
Although information missing there is, it  
 is possible to recover these vessels from the grayscale image.
 We also notice that the vessel share key features with line segment.
 }
\label{fig:grayscale-inspiration}
\end{figure}

%
The proposed approach leverages the DCNN's strong power in feature 
learning. We achieve this labeling-free approach via generating samples
with prior knowledge and training the networks on the data set.
The prior knowledge must be strong enough to guarantee desirable performance.
This idea is inspired by the observation that blood vessels are 
clearly delineated in grayscale image,
as \figurename{\ref{fig:grayscale-inspiration}} illustrates,
where we take an image from \emph{DRIVE} 
data set \cite{staal:2004-855}
 as an example. 
 
 Based on this observation, we assume that 
 \emph{if a deep neural networks model can distinguish
 	line segment from noise, it would be compatible with blood vessel}.
 	We  hypothesize
 that \emph{with the great power of DCNN in feature 
 learning, one can segment these vessels in grayscale images
 by training the nets on artificial data set automatically generated from
  prior knowledge}. That is, no labeling is needed.
To illustrate our idea,
we  use simple line segments 
to construct the samples, which is
followed by a post-processing aimed
 to provide more prior knowledge;
 and then  a naive \emph{DCNN} model  is built and trained 
 to test our approach.


Generating samples functions as a key component, in our approach, 
since the images in our training data set are artificial (they are generated by 
computing). We work out this stage with two steps. First is to generate 
raw images, and then further processing is invoked 
to approximate real-world images more closely.
In this section, we demonstrate the processes of generating raw images and
improving the data set, respectively.

\subsection{Raw Image Generation}
In this phase, we construct images 
with simple line segments on
empty background, based on the 
following rules we developed.

\emph{Rule 1-1}
: Most of the line segments' nodes should be 
connected one after another.
In real-world images, as we observe,
the graph of the vessels is likely to contain many cliques.
And we consider it one of the features 
the system needs to learn. That is,
they are not being simply imposed, but of 
nontrivial structures.

\emph{Rule 1-2}
: The gray level of each line segment 
is expected to differ from others.
This provides an alternative way to
simulate the variation of illumination.
While the intensity of the pixels,
apparently, may vary locally in
real-world images, it gets complicated to 
make a pixel-wise assignment.
We simplify this process by using 
line-wise assignment instead, regarding to our hypothesis.

\emph{Rule 1-3}:
Line segment's length should be
able to vary in a predefined range.
This rule is based on the observation
that the blood vessels are not strictly
straight, but zigzag in local. 
We make it a key component to bridge 
the gap between the simulation and 
the reality-based image, together 
with the direction variation (will be 
addressed in \emph{Rule \mbox{1-4}}).

\begin{figure}
\centering
\includegraphics[scale=0.44]{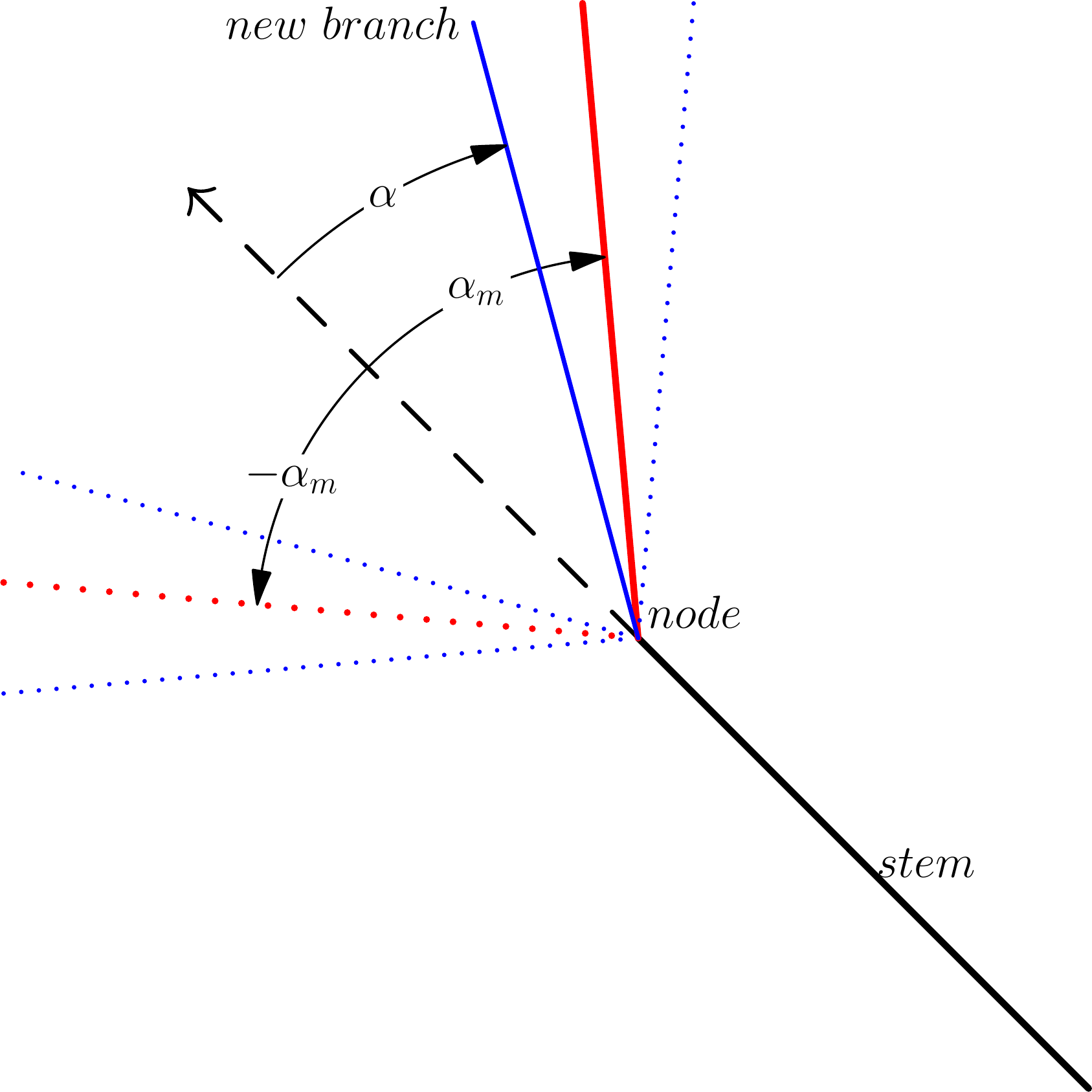}
\caption{Illustration of \emph{Rule 1-4}. There are two
candidate angles ($\pm\alpha_m$), and we randomly choose one ($+\alpha_m$, in this case)
as the {mean} of $\alpha$ to draw a sample.
 As a result, the {\bf{new~branch}} (blue, in this case, the solid one)
 has a symmetry distribution (i.e. {normal distribution}) centered at the red line
 (bold, in this case, the solid one).}
\label{fig:illu-rule4}
\end{figure}

\emph{Rule 1-4}:
The structure is encouraged to spread over the image while
keep sparse.
This rule focuses on the layout of the blood vessels.
We notice that the vessels tend to cover the whole 
image and each vessel is distinct as they are clearly
separated from each other. Let $\alpha$ denote the angle 
between a new branch (line segment) and its stem's direction,
to take account of this prior information, 
we set a predefined pair of angles (i.e. $\pm\alpha_m$ ) as candidate means for 
$\alpha$, as illustrated in \figurename{\ref{fig:illu-rule4}} where 
{normal distribution}  has been used to generate a sample of $\alpha$
(i.e. $\alpha\sim\mathcal{N}(\pm\alpha_m,\sigma_{\alpha})$).

To facilitate the construction, more details should be 
specified.
In \emph{rule 1}, we limit the number of the
\emph{nodes} (points where \emph{new branches} begin, see \figurename{\ref{fig:illu-rule4}})
, which 
incorporates with \emph{rule \mbox{1-4}} in terms of keeping sparse.
In \emph{rule \mbox{1-2}}, a random value is assigned to a line segment as 
its gray level. In \emph{rule \mbox{1-3}}, we define 
the \emph{{mean}} of the length
to facilitate the generation process. 
Besides that, we keep the line segments in a specified circular region; 
when a new branch tends to break out, we reject it and draw a new one.
This trick tends to encourage the diversity of the structures, which 
approximates the zigzag feature more precisely.

\begin{figure}

\centering
\begin{subfigure}[b]{0.23\textwidth}
\centering
\includegraphics[scale=0.23]{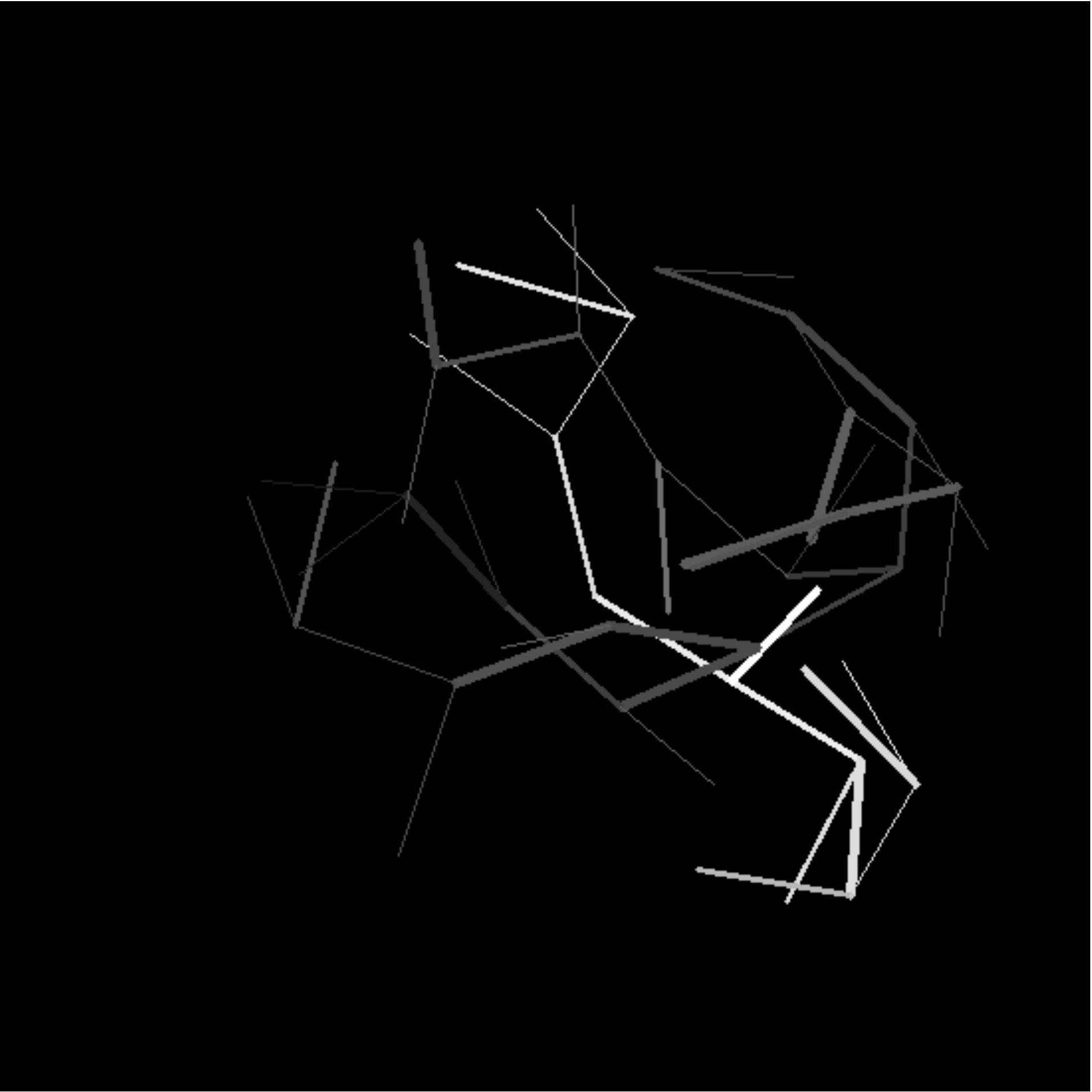}
\end{subfigure}
~
\begin{subfigure}[b]{0.23\textwidth}
\centering
\includegraphics[scale=0.23]
{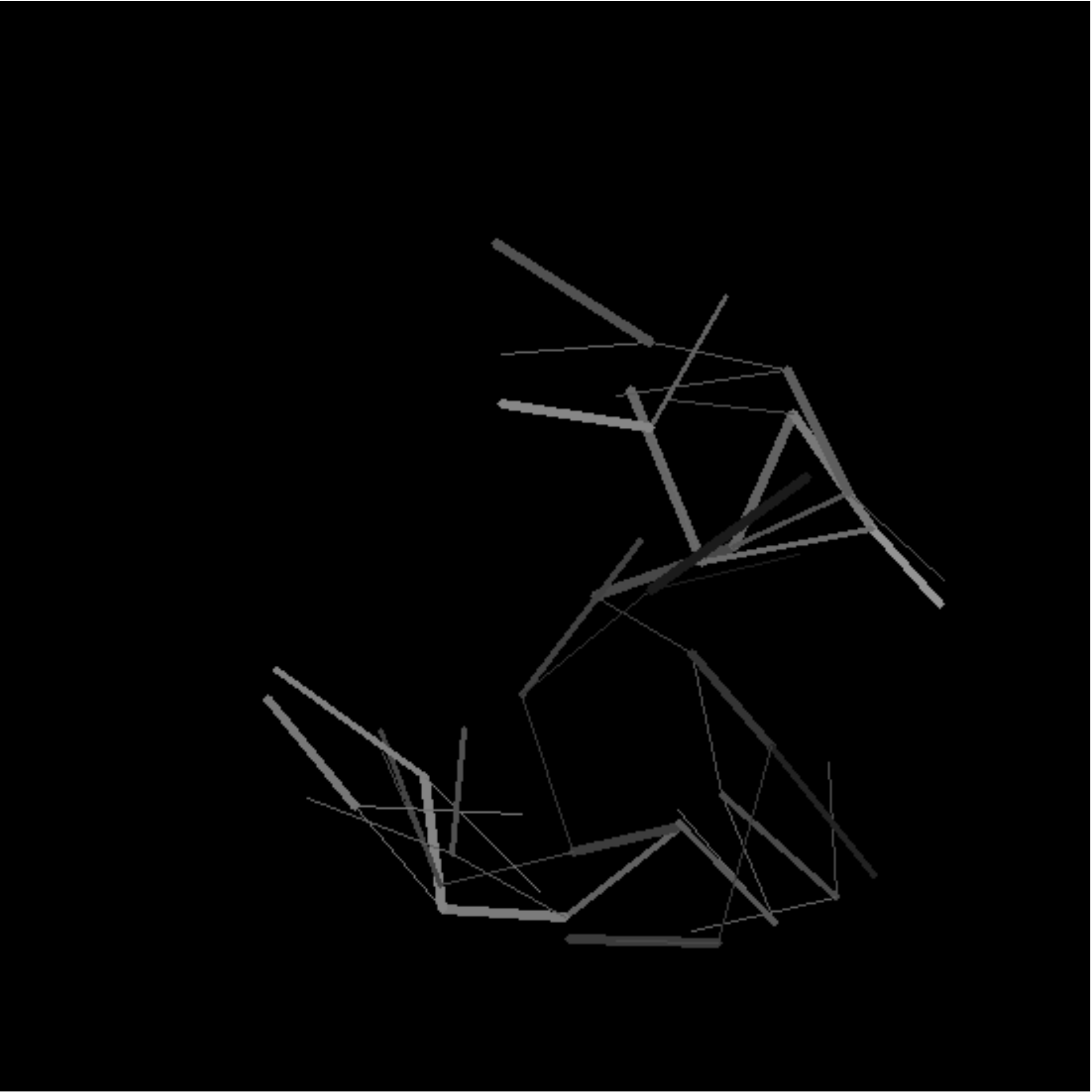}
\end{subfigure}
\caption{Generated raw images based on our rules.
They are constructed with only simple lines, but come with
distinct features abstracted from the real-world objects, which makes 
it possible to use them as training samples. 
See \emph{rule 1-1}$\sim$\emph{1-4} and
 \emph{Algorithm \ref{algo:algorithm1}} for more details.
}
\label{fig:rawImages}
\end{figure}

\begin{algorithm}
\SetAlgoLined
//construct image with simple lines

{\bf Parameters}: {$\alpha_m,N,\sigma_L,L_m,max_{chd}
,\sigma_\alpha$}

$img\longleftarrow \bm{0}$

\While{$node~number~<~N$}
{
	\While{$chd_{[nodeIdx]}~<~max_{chd}$}
	{
		$length \sim \mathcal{N}(L_m,\sigma_L)$
		
		$\alpha \sim \mathcal{N}(\pm\alpha_m,\sigma_\alpha)$
		
		$(x,y) \longleftarrow ~genPoint(P_{[nodeIdx]},\alpha,len)$
		
		\If{$(x,y)$ is out if the circle}
		{
			continue
		}
		$grayscale    \sim  uniform~distribution$

		draw line from $P_{[nodeIdx]}$ to $(x,y)$ in $img$
		
		record $(x,y)$ in $P$
		
		update variables
	}
	
	\
\
}
return $img$
\caption{Raw Image Generation}
\label{algo:algorithm1}
\end{algorithm}

With these rules, we developed \emph{Algorithm \ref{algo:algorithm1}};
simply, we use uniform distribution and normal distribution
in the algorithm.
Two of the generated images are shown in \figurename{\ref{fig:rawImages}},
and we notice that the labels are automatically obtained.

\subsection{Improvement of the Samples}
Typically, the process of training on large data set is to combat the noise
existing in the data set. And, in this work, 
our samples come with two kinds of noise; 
the first comes from the 
approximation of the vessels' structure;
 the other, similar to the
real images, arises from the background.

Actually, in real-world images, noise in background comes as 
a major problem hampering the segmentation.
That is,  if the background of
\figurename{\ref{fig:grayscale-inspiration}} 
          was as clear as  
that of \figurename{\ref{fig:rawImages}}, 
we would segment these vessels in straightforward ways.

In this stage, we tend to add artificial noise
to the background for providing sufficient prior information.
We construct the noise by combining two types of 
artificial noise, namely, global noise and local noise.
We also make rules to develop our algorithm for
improving our samples, with respect to the two types of 
noise.

\emph{Rule 2-1}:
There is global background noise in sample.
The global background noise is designed to 
fully utilize the whole pixels to avoid 
trivial solution. 
In our implementation, we first add one random
value to the image at each pixel; and then generate
\emph{Gaussian noise} for each pixel
using identical parameter sets.

\emph{Rule 2-2}:
Local noise is necessary.
Apparently, uniform Gaussian noise is insufficient
to deliver our observation-based prior knowledge.
To accomplish that, we consider introducing 
specific noise to the randomly selected patches (with fixed size).
The noise should be capable of varying smoothly,
which is to capture the feature of illumination variation.
Under this goal, we use sine wave to generate these 
noise.
For the reason of sample's diversity and convergence of 
training process, we randomly draw a value as the \emph{phase} 
while keep both
\emph{amplitude} and \emph{frequency} fixed.
The frequency's setting is adapted to the patch's
size to guarantee sufficient intensity variation.

\begin{figure}
\centering
\begin{subfigure}[b]{0.23\textwidth}
\centering
\includegraphics[scale=0.23]{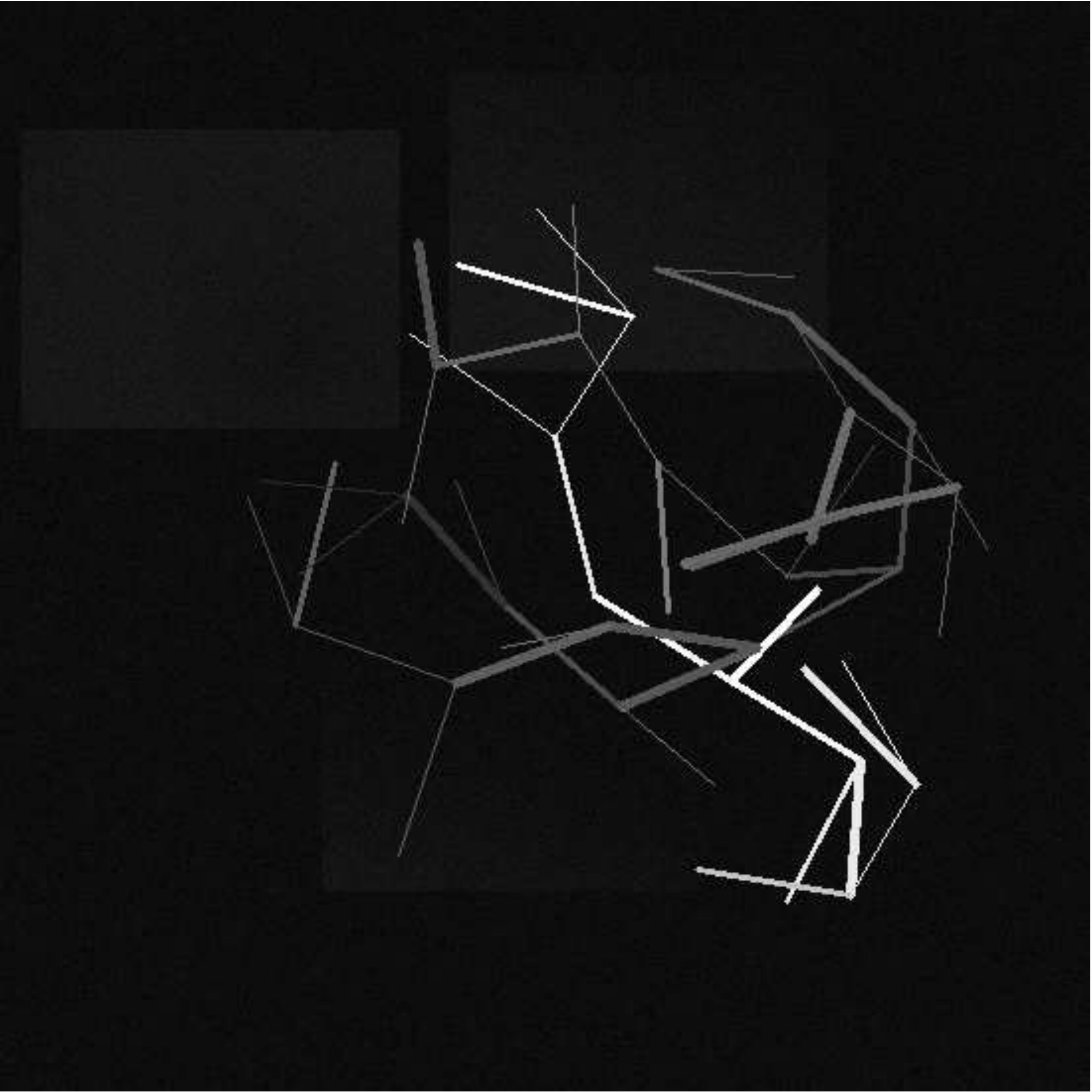}
\end{subfigure}
~
\begin{subfigure}[b]{0.23\textwidth}
\centering
\includegraphics[scale=0.23]{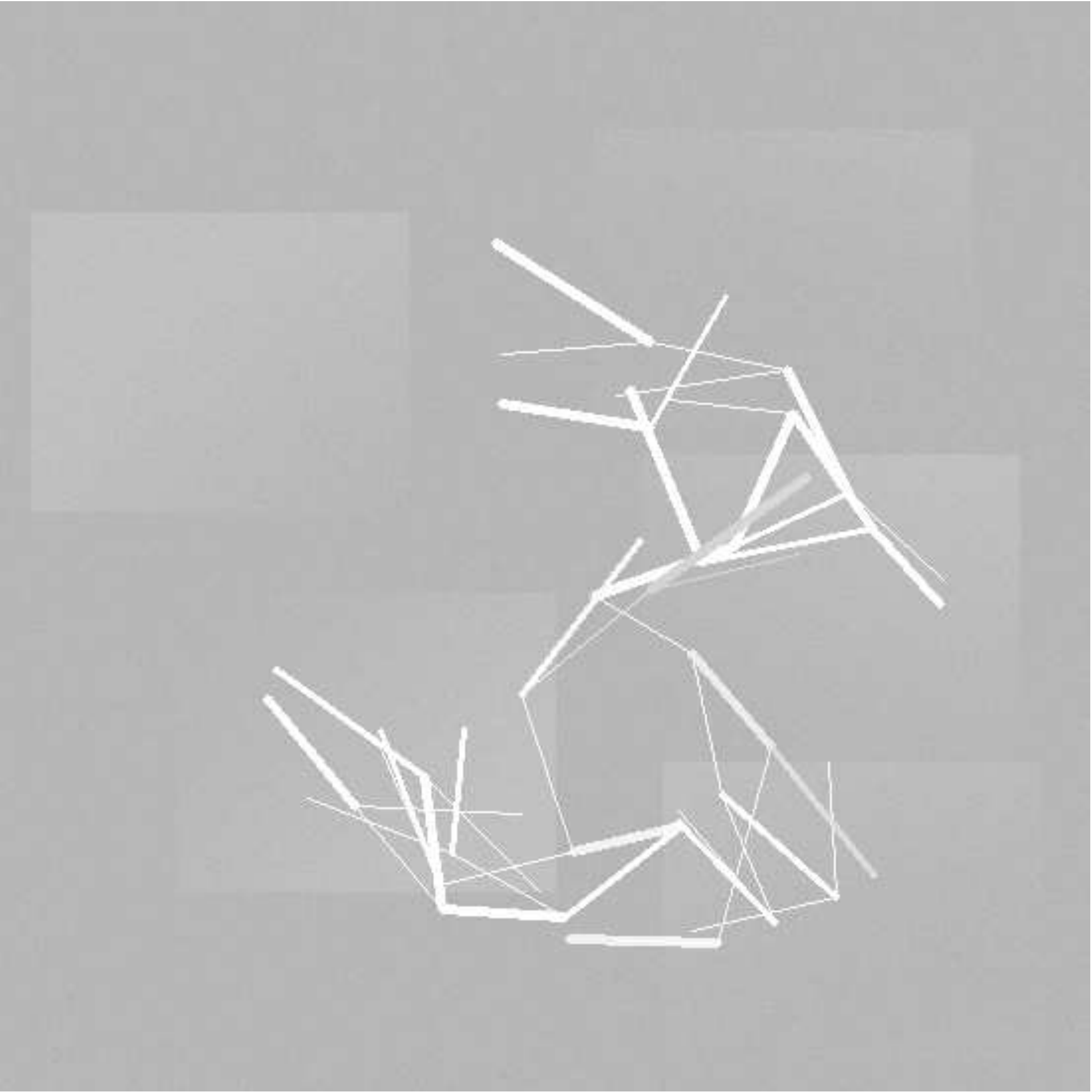}
\end{subfigure}
\caption{Generated samples based on our rules.
Background noise has been added to avoid trivial solutions.
In this case, we set $N_{max}=5$. 
See \emph{rule 2-1,2-2} and \emph{Algorithm \ref{algo:algorithm2}}
for more details.
}
\label{fig:samples}
\end{figure}

The philosophy behind pouring noise,  
to the raw images is to avoid trivial solution (over-fitting),
similar to the novel work in \cite{vincent2010stacked}.
We expect the nets to preserve and enhance the structures
which share key features (e.g. being thin and continuous)
with the vessels in real-world images, 
while suppress the noise.  
From this perspective, in the proposed approach,
the networks works as a filter which boosts the \emph{SNR}
in the output.

We derive \emph{Algorithm \ref{algo:algorithm2}}
to improve our samples based on \emph{rule 2-1} and \emph{rule 2-2}.
Some finally generated samples are shown in \figurename{\ref{fig:samples}}.

\begin{algorithm}
\SetAlgoLined
//add background noise to a sample

{\bf Parameters}: {$m_{noise},\sigma_{noise},N_{max},\omega,A$}

// $img$ is initially generated by \emph{Algorithm} \ref{algo:algorithm1}

// global noise

$b \sim  Uniform~distribution$ //a scalar

$n_g\sim \mathcal{N}(m_{noise},\sigma^2_{noise})$ //$i.i.d.$, same size as $img$'s

select $n(n<N_{max})$ patches  // no overlap 

// local noise

\For{ each patch \emph{P}}
{

	\For{each pixel $(x,y)$ in \emph{P}}
	{
		$\alpha\sim  Uniform~distribution$
		
		$img_{[x,y]}\longleftarrow img_{[x,y]}+A\sin(\omega f(x,y)+\alpha)$		
		// $f(\cdot,\cdot)$ is a distance function
	}
}
$img\longleftarrow img+n_g+b$

return $img$

\caption{Sample Generation }
\label{algo:algorithm2}
\end{algorithm}

\section{Experiments}
To test our hypothesis, in this section, we
build a naive deep neural networks model and trained it on the data set, 
generated by the algorithms,
 and then tested it on \emph{DRIVE} \cite{staal:2004-855}
  and \emph{STARE}  \cite{hoover2000locating} data set.
   To explore the importance of prior knowledge,
   we used two types of samples to train the model respectively.
   
\subsection{Datasets}

We evaluated our approach on two benchmark databases: \textit{{DRIVE}},
and \textit{STARE} databases.

\subsubsection{\textit{{DRIVE}} Database}
The DRIVE \cite{staal:2004-855} (Digital Retinal Images for Vessel Extraction) database
has been established to enable comparative 
studies on segmentation of blood vessels in retinal images.
The images were acquired using a Canon 
CR5 non-mydriatic 3CCD camera with a 45 degree field of view (FOV).
 Each image was captured using 8 bits per color plane at 768 by 584 pixels. 
 The FOV of each image is circular with a diameter of approximately 540 pixels.
  For this database, the images have been cropped around the FOV. For each image, 
  a mask image is provided that delineates the FOV.

  The set of 40 images has been divided into a training and a test set, both containing 20 images.
   For the training images, a single manual segmentation of the vasculature is available.
   For the test cases, two manual segmentations are available;
    one is used as gold standard, the other one can be used to compare computer generated segmentations with those of an independent human observer.
 All human observers that manually segmented the vasculature were instructed and trained by an experienced ophthalmologist. 
     They were asked to mark all pixels for which they were for at least $70\%$ 
     certain that they were vessel. 
     We used the first observer's outputs as the ground truth.
     We only used the test set since our approach only took artificial images
     for training.

\subsubsection{\textit{{STARE}} Database}
The STARE \cite{hoover2000locating}
(Structured Analysis of the Retina) database contains 20 images for blood vessel segmentation,

The slides are obtained by TopCon TRV-50
fundus camera  at 35$\deg$
field of view. Each slide  was digitized
to produce a $605\times700$ pixel image, 24 bits per pixel (standard RGB).
 Ten of the images are of patients with no pathology
(normals). Ten of the images contain pathology that obscures or
confuses the blood vessel appearance in varying portions of the
image (abnormals). The database contains two sets of manual
segmentations prepared by two observers, and the former one
is considered as the ground truth \cite{liskowski2016segmenting}.

%
\subsection{Experiments Settings }
\subsubsection{Deep Neural Networks Model}

\begin{figure*}
  \centering
  \includegraphics[height=0.3\textheight]{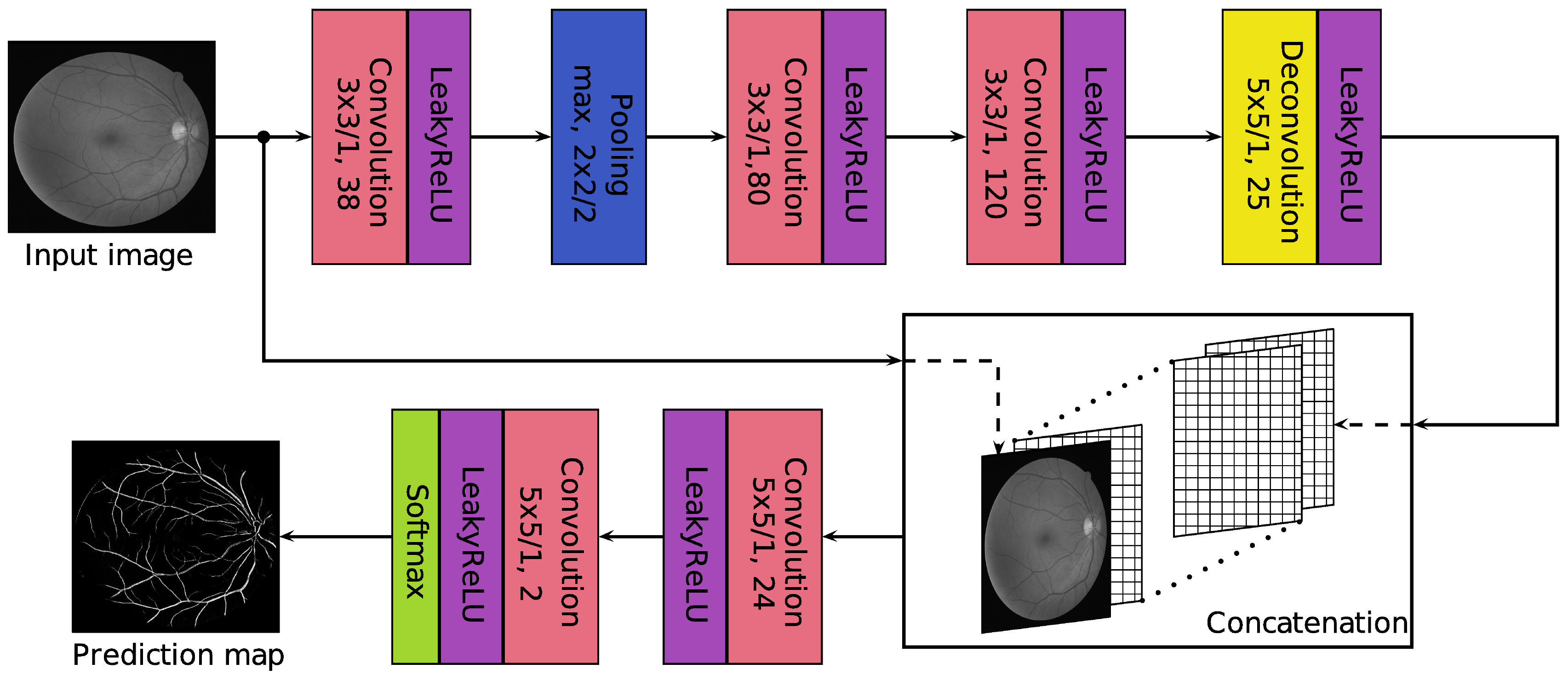}
  \caption{Our networks model used for segmentation.
  We use this model to test our hypothesis. See \emph{Experiments} for 
  more details.}
  \label{fig:nets}
\end{figure*}
 The networks (see \figurename{\ref{fig:nets}}) are built without fully connected layer,
 which makes it portable to images of different sizes. This allows us to
  use smaller images as training samples to accelerate our training process. And 
the loss layer is a pixel-wise \emph{softmax} layer (used for pixel-wise classification).
We build a \emph{concatenation} layer to follow the philosophy that
high-level perception guides the work in lower levels. 
In the concatenation layer and the softmax layer,
cropping has been used to handle the differences in size of the inputs. 
The implementation 
is based on {MXNet} \cite{Chen2015MXNet}\footnote{http://mxnet.io}.
The experiments were conducted on hardware configuration: Intel
Core i3-530 CPU with single NVIDIA GTX 760 graphics
card, software configuration: Ubuntu 14.04, Python 2.7.

The networks  have been trained  
with default hyperparameter settings, and \emph{Batch Normalization} 
\cite{ioffe2015batch} has been applied.

\subsubsection{Training Datasets}
We built two types of training datasets to explore
the differences caused by different prior knowledge.
Images (e.g. \figurename{\ref{fig:experiment-thick-sample-drive}}) in \textit{dataset$\#1$} are constructed with 
wider and more distinctive lines, while images
 (e.g. \figurename{\ref{fig:experiment-thin-sample-drive}})
in \textit{dataset$\#2$} only contain lines of single type,
and the noise in background makes them difficult to detect.

\subsection{Performance Measurements}

\begin{table}
\begin{center}
\caption{Performance metrics for retinal vessel segmentation}
\label{tab:performance-metrics}
\begin{tabular}{cc}
\hline
Performance measures & Description\\
\hline
Sensitivity(Sn) & TP/(TP+FN)\\
Specificity(Sp) & TN/(TN+FP)\\
Accuracy (Acc)  & (TP+TN)/(TP+FP+TN+FN)\\
True positive rate (TPR)  & TP/(TP+FN)\\
False positive rate (FPR)  & FP/(FP+TN)\\
AUC    &  Area under the ROC curve\\
\hline
\end{tabular}
\end{center}
\end{table}

In this work, we treat the segmentation as a binary 
classification task. That is, each pixel will be either classified as
\emph{blood vessel} or \emph{background}.
 For this reason, we evaluated our  networks' performance 
 in terms of \emph{{AUC}},
 \emph{{Sensitivity}}, \emph{{Specificity}} and \emph{{Acc}} (accuracy).
  Their definitions are described in \tablename{\ref{tab:performance-metrics}}
  where TP (true positive) means the number of pixels correctly classified into 
  vessels, {FP} (false positive) is the number of pixels misclassified into vessels,
  {TN} (true negative) is the number of pixels correctly classified into background,
  and {FN} (false negative) is the number of pixels misclassified into background.
  ROC (receiver operation characteristic) curve is obtained by plotting 
  the {TPR} against the {FPR} at various threshold settings.
  In our experiments,
\mbox{\textit{{scikit-learn}}}\footnote{http://scikit-learn.org
   } has been utilized to 
  calculate {{AUC}} and {{Acc}}.

\subsection{Evaluations}
In this subsection, we evaluate our nets, and explore
the effects caused by different training datasets.
The results show that:
\begin{itemize}
\item our networks achieve competitive performance among the 
state-of-the-art methods;
\item prior knowledge is essential for our proposed approach.
\end{itemize}

\begin{figure}
\centering
\begin{subfigure}[b]{0.23\textwidth}
\centering
\includegraphics[scale=0.23]{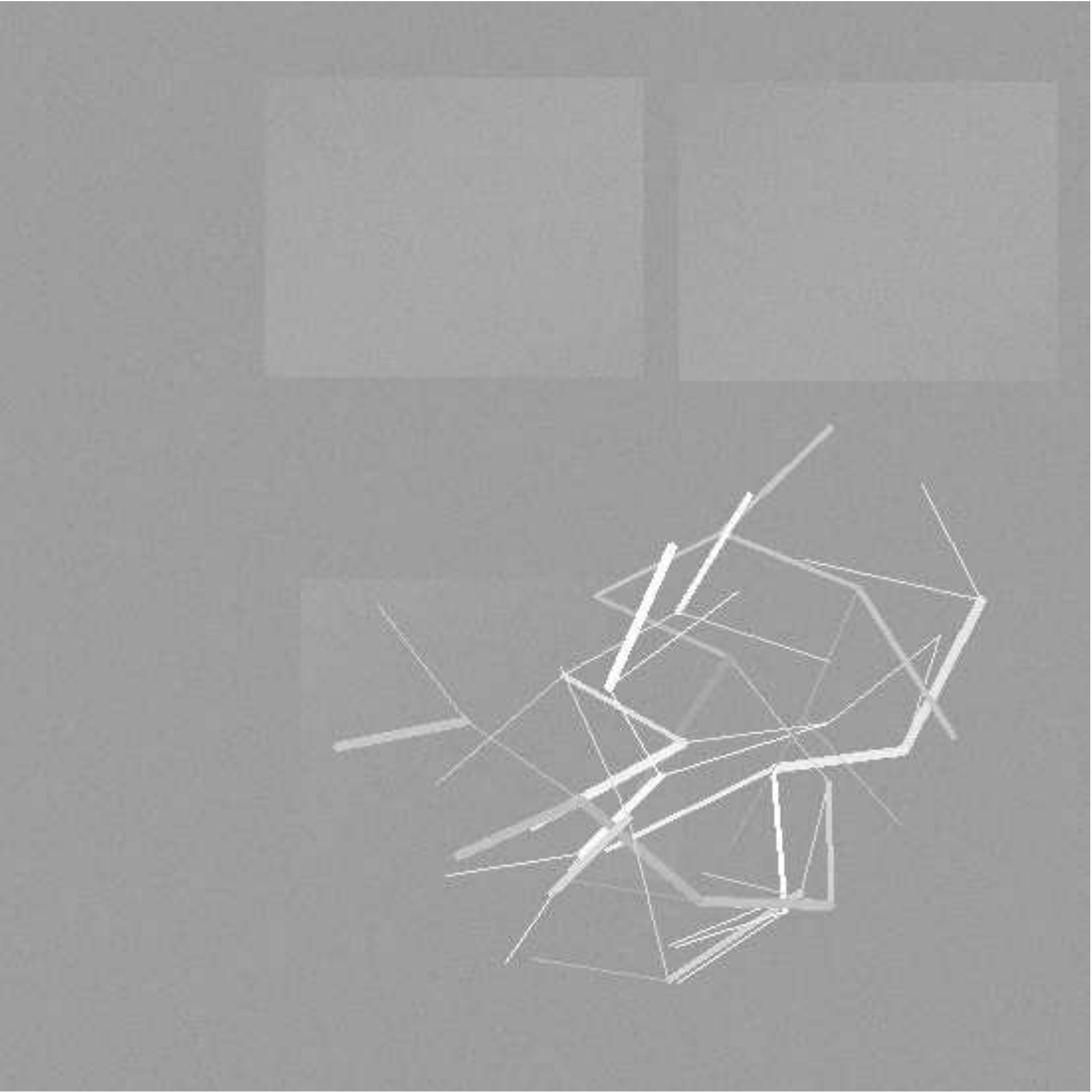}
\end{subfigure}
 ~
\begin{subfigure}[b]{0.23\textwidth}
\centering
\includegraphics[scale=0.23]{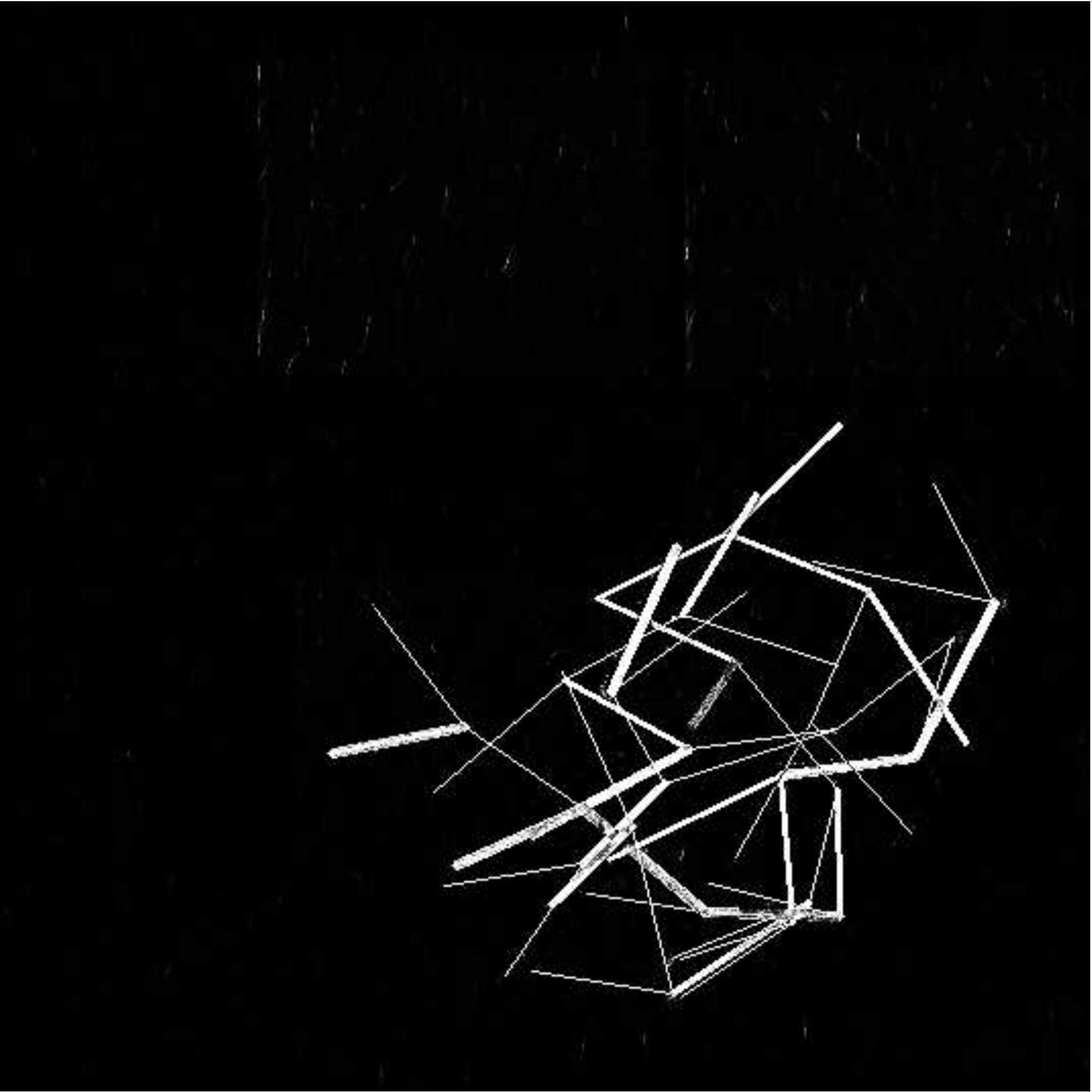}
\end{subfigure}
\caption{Prediction map (right) on sample (left).
   Structures with desired features are preserved. Because of the
   convolution operations, they have different sizes. The nets are 
   trained on \textit{dataset$\#2$}.}
\label{fig:predictOnSample}
\end{figure}

We used some of the samples to test our nets. The  results
 are shown in \figurename{\ref{fig:predictOnSample}}.
  As the result shows, the model has
 been able to distinguish line segments from the noise.

\begin{figure}
 \centering
 \begin{subfigure}[b]{0.2\textwidth}
 \centering
 \hfil
 \includegraphics[scale=.2]{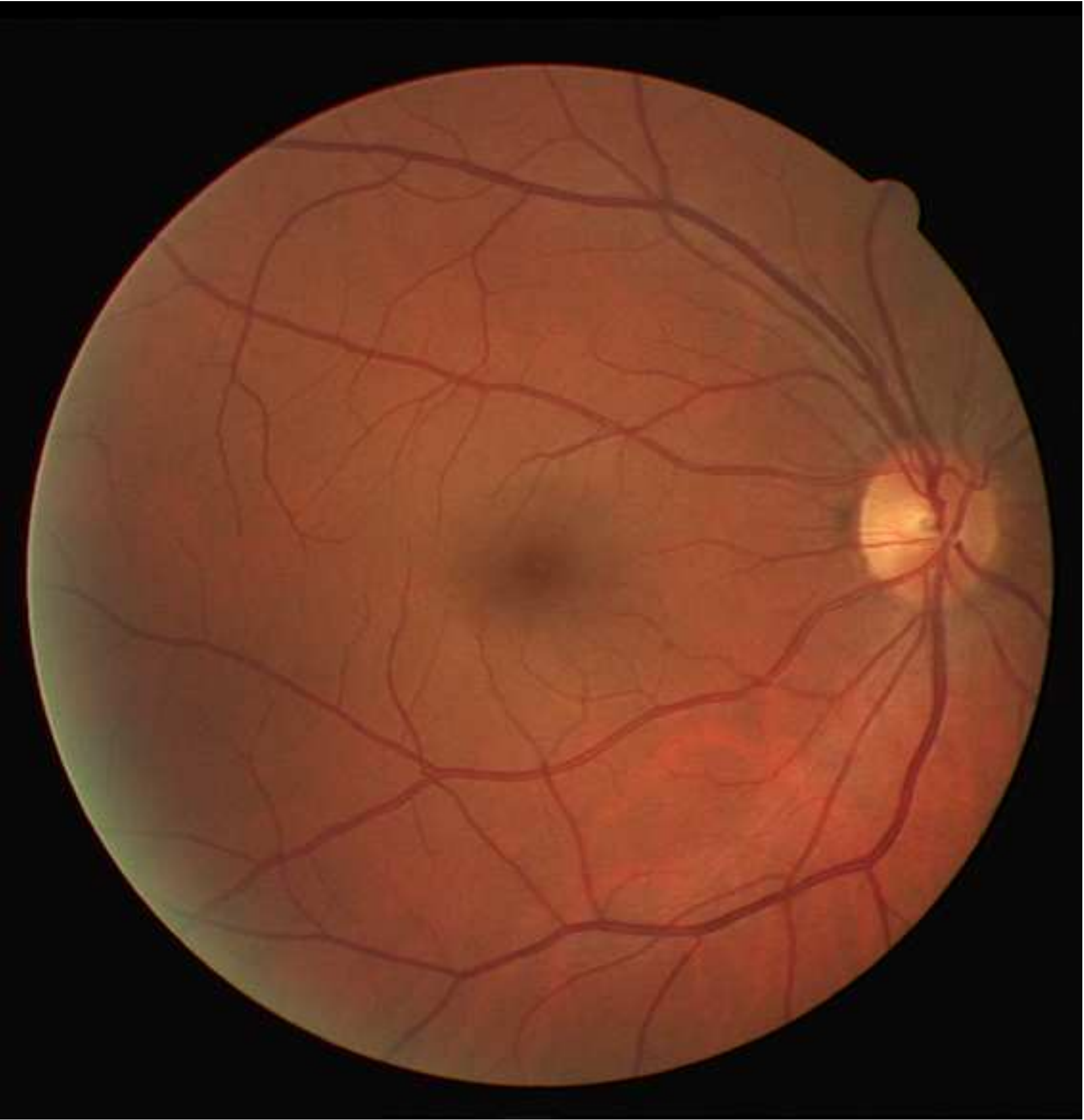}
 \caption{} 
 \label{fig:experiment-real-drive}
 \end{subfigure}
    \quad
 \begin{subfigure}[b]{0.2\textwidth}
    \centering
    \hfil
    \includegraphics[scale=0.2]{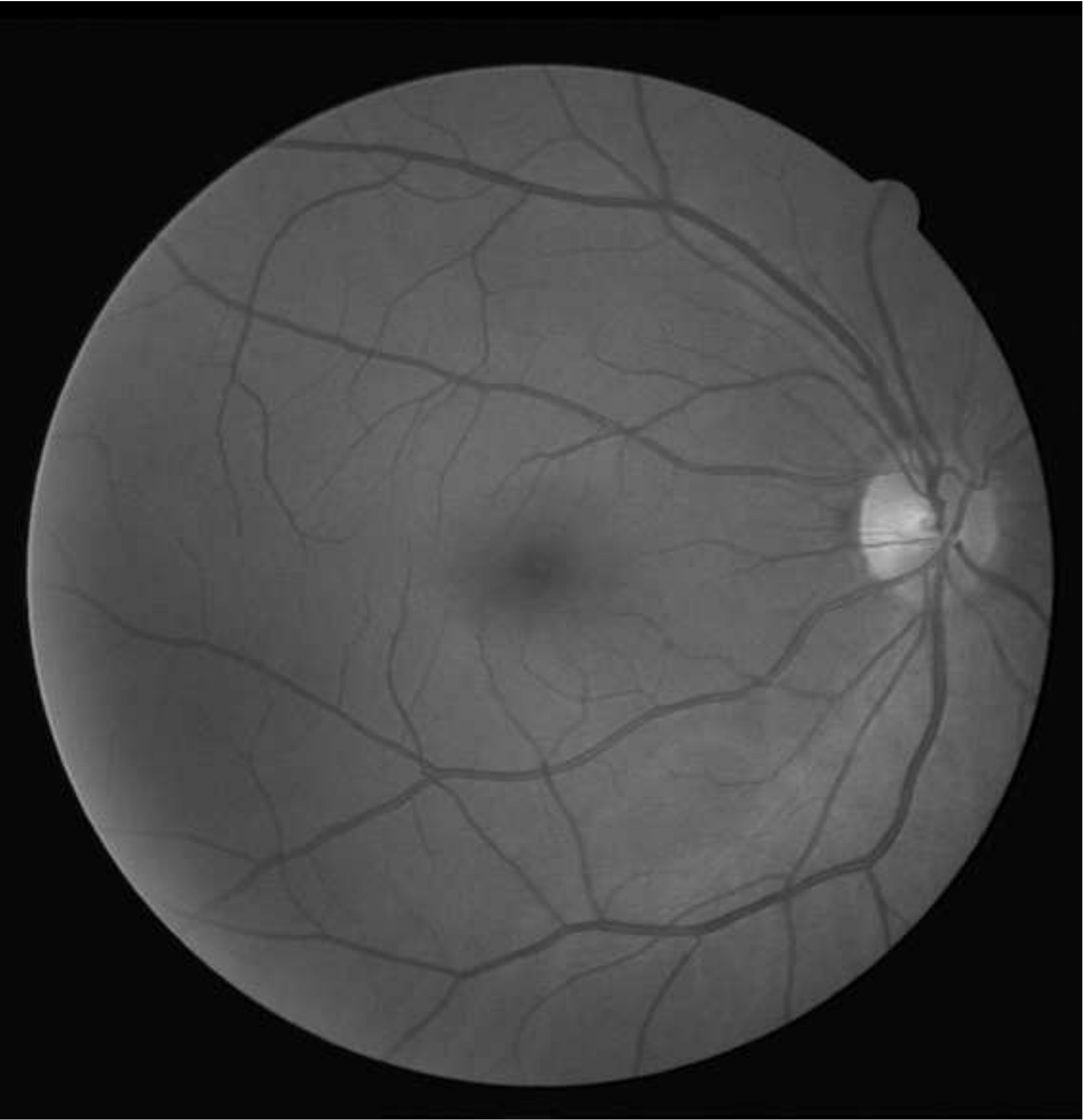}
    \caption{} 
    \label{fig:experiment-gray-drive}
 \end{subfigure}
    \\
 \begin{subfigure}[b]{0.2\textwidth}
 \centering
 \includegraphics[scale=0.2]{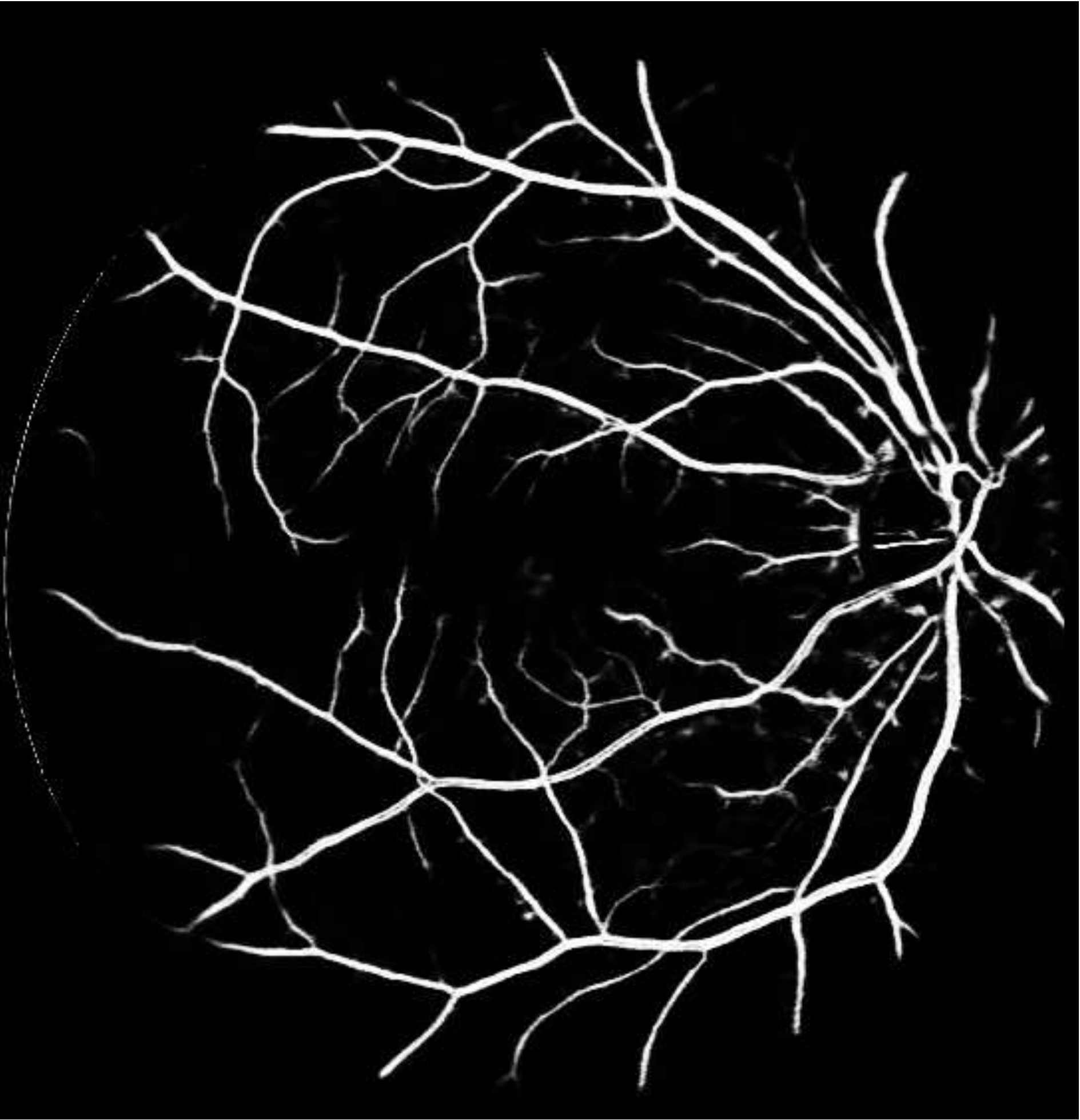} 
 \caption{} 
 \label{fig:experiment-prob-drive}
 \end{subfigure}
 \quad
 \begin{subfigure}[b]{0.2\textwidth}
 \centering
 \includegraphics[scale=0.2]{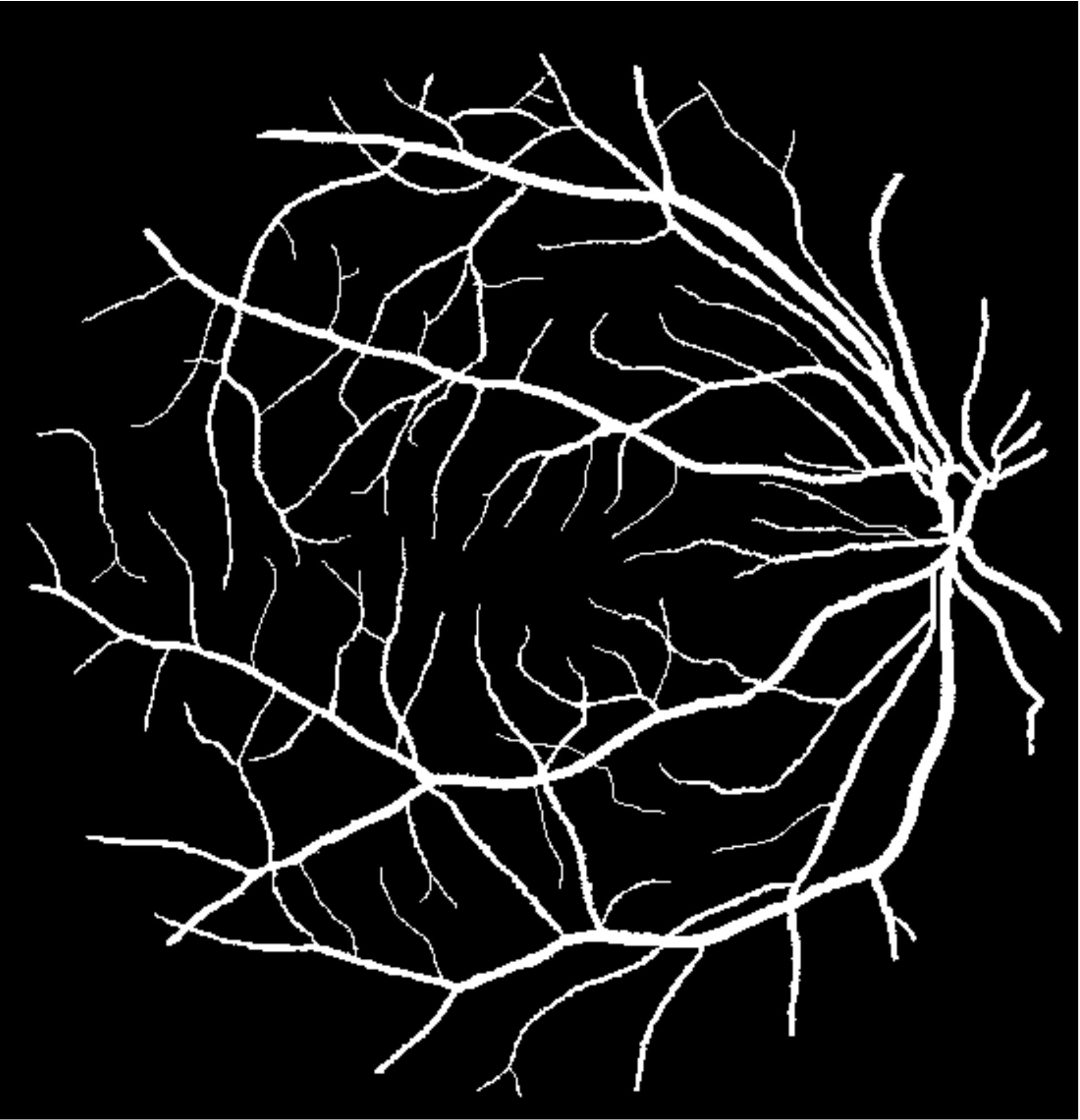}
 \caption{} 
 \label{fig:experiment-gdt-drive}
 \end{subfigure}
 \hfil
 \caption{Best result (according to AUC) of the trained networks on \textit{{DRIVE}} test 
    data set (using \textit{dataset$\#1$}). (a) 
    Image from \textit{{DRIVE}} test data set (b) Grayscale version (c) Our prediction map 
    (d) Ground truth.}
 \label{fig:predictOnReal}
\end{figure}

 Then, we evaluated the networks on \textit{DRIVE}  
 and \textit{STARE} data sets.
 For a given image (e.g. \figurename{\ref{fig:experiment-real-drive})
 , we used its grayscale version (\figurename{\ref{fig:experiment-gray-drive}}) 
 to feed the networks, before that, an image inversion was required 
 because of the difference between the training sample and the real image
  (see \figurename{\ref{fig:samples}} 
 and \figurename{\ref{fig:grayscale-inspiration}}).
The results are shown in \figurename{\ref{fig:predictOnReal}} 
and \tablename{\ref{tab:performance-drive-stare}}.

 The results of \figurename{\ref{fig:predictOnSample}}
 and \figurename{\ref{fig:predictOnReal}} 
  imply the mechanism of the neural networks.
 That is, the deep neural networks give strong responses to line-like structures,
  while ignore others. This is the feature we are working for; and, as the 
  results show, it yields a system compatible with vessel's structure.

\begin{figure*}
  \centering
  \begin{subfigure}[b]{0.3\textwidth}
  \centering
  \hfil
  \includegraphics[height=.2\textheight]{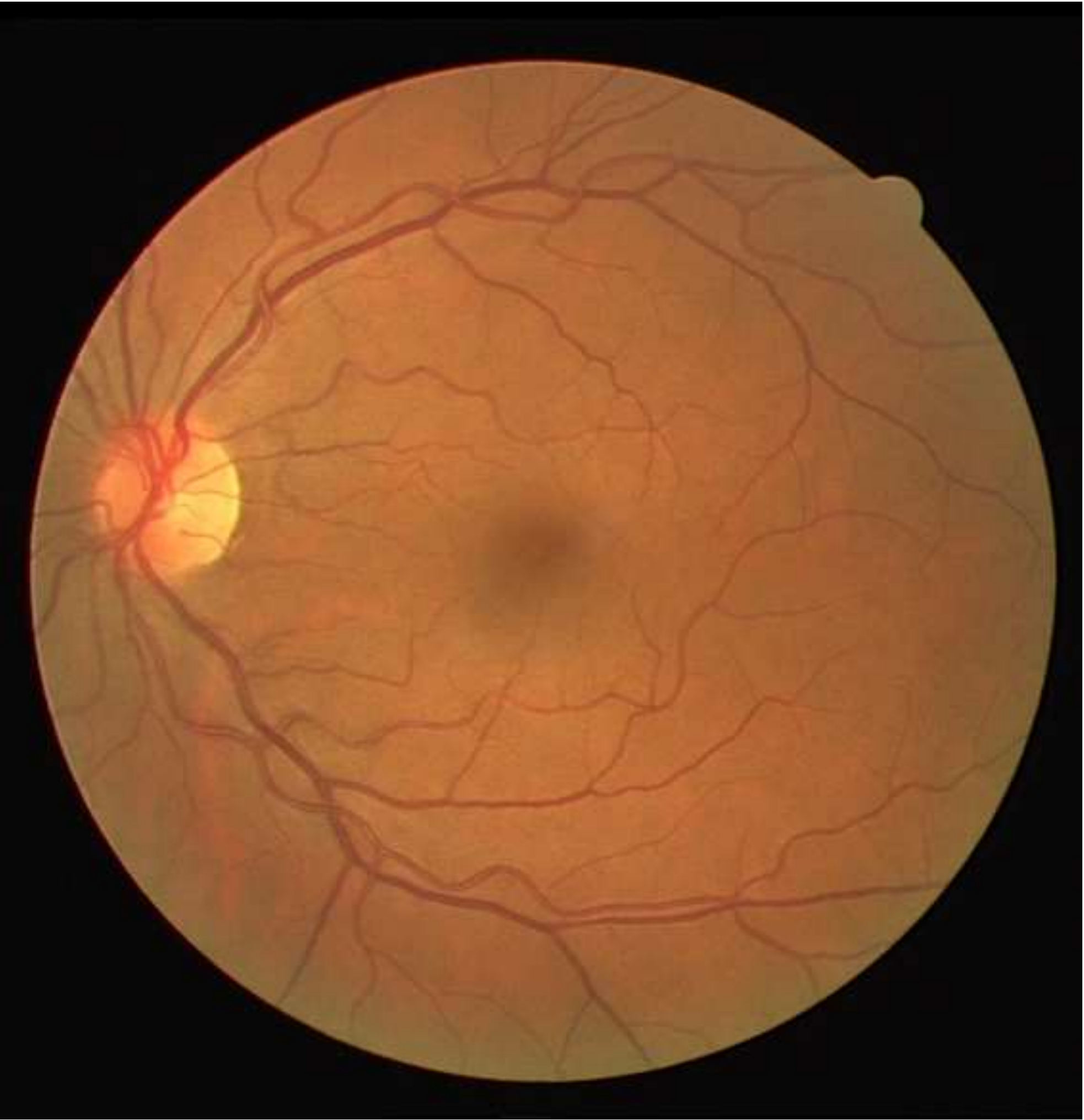}  
  \caption{} 
  \label{fig:experiment-01_test-drive}
  \end{subfigure}
    ~
  \begin{subfigure}[b]{0.3\textwidth}
  \centering
  \includegraphics[height=.2\textheight]{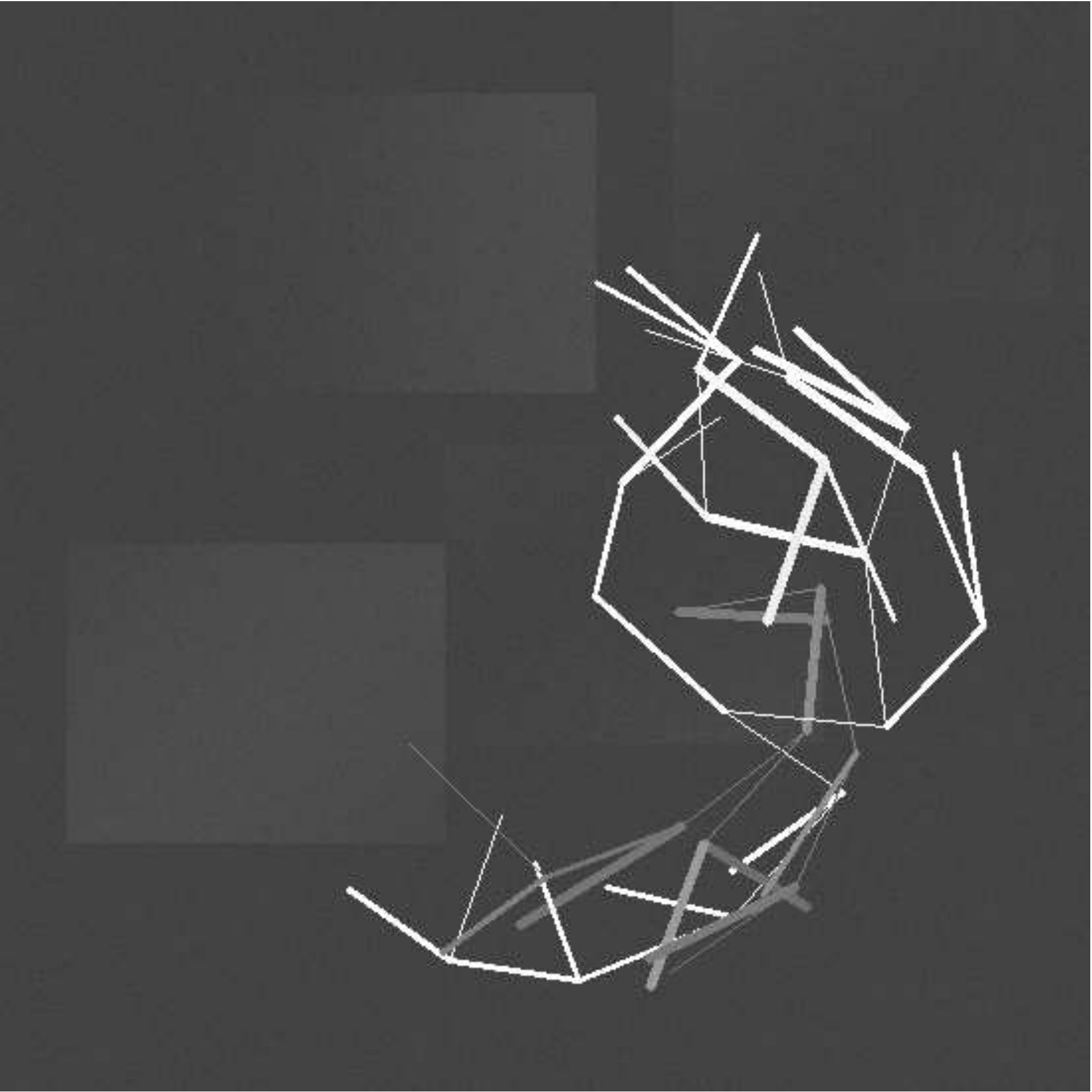} 
  \caption{} 
  \label{fig:experiment-thick-sample-drive}
  \end{subfigure}
  ~
  \begin{subfigure}[b]{0.3\textwidth}
  \centering
  \includegraphics[height=.2\textheight]{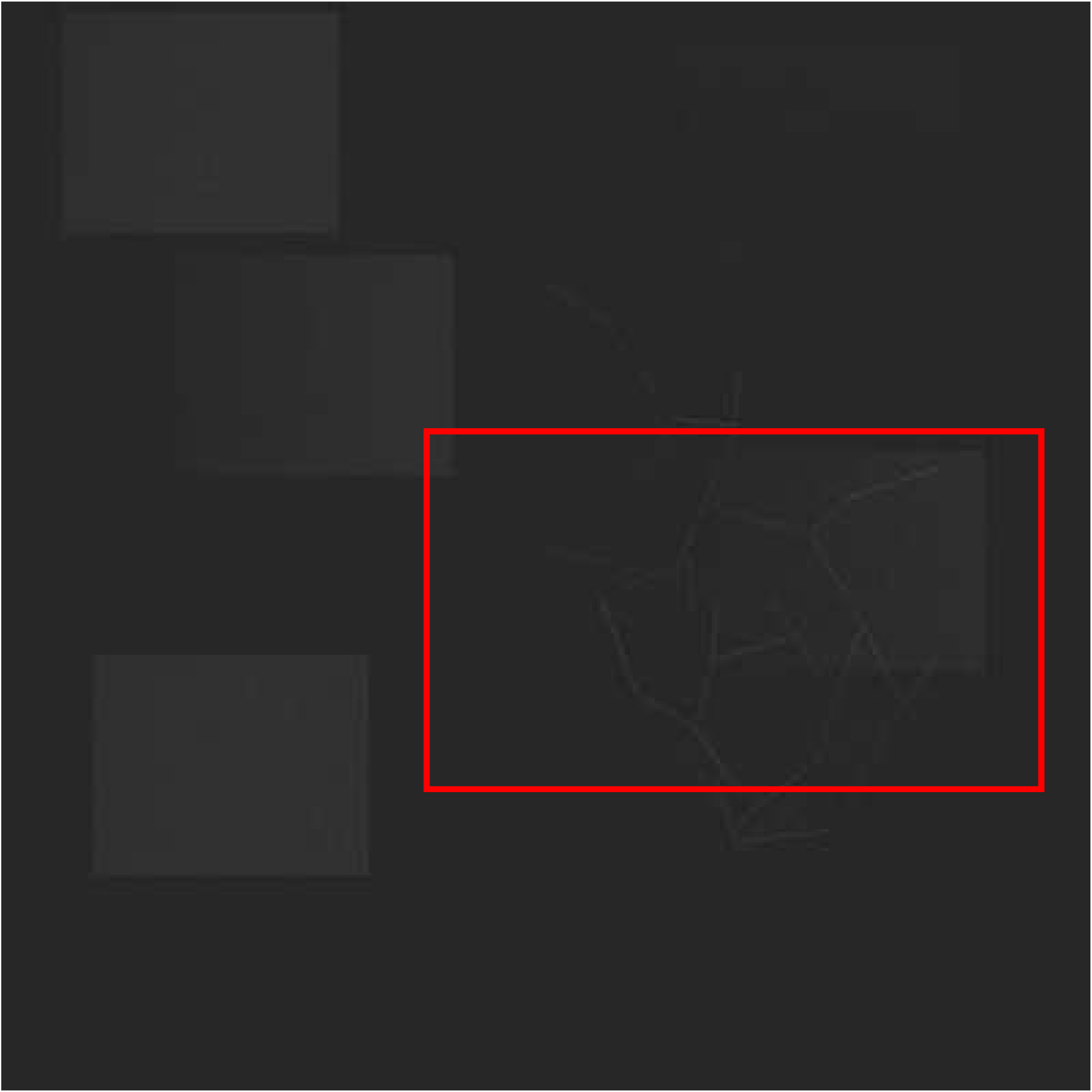} 
  \caption{} 
  \label{fig:experiment-thin-sample-drive}
  \end{subfigure}
  \\
  \begin{subfigure}[b]{0.3\textwidth}
  \centering
  \hfil
  \includegraphics[height=.2\textheight]{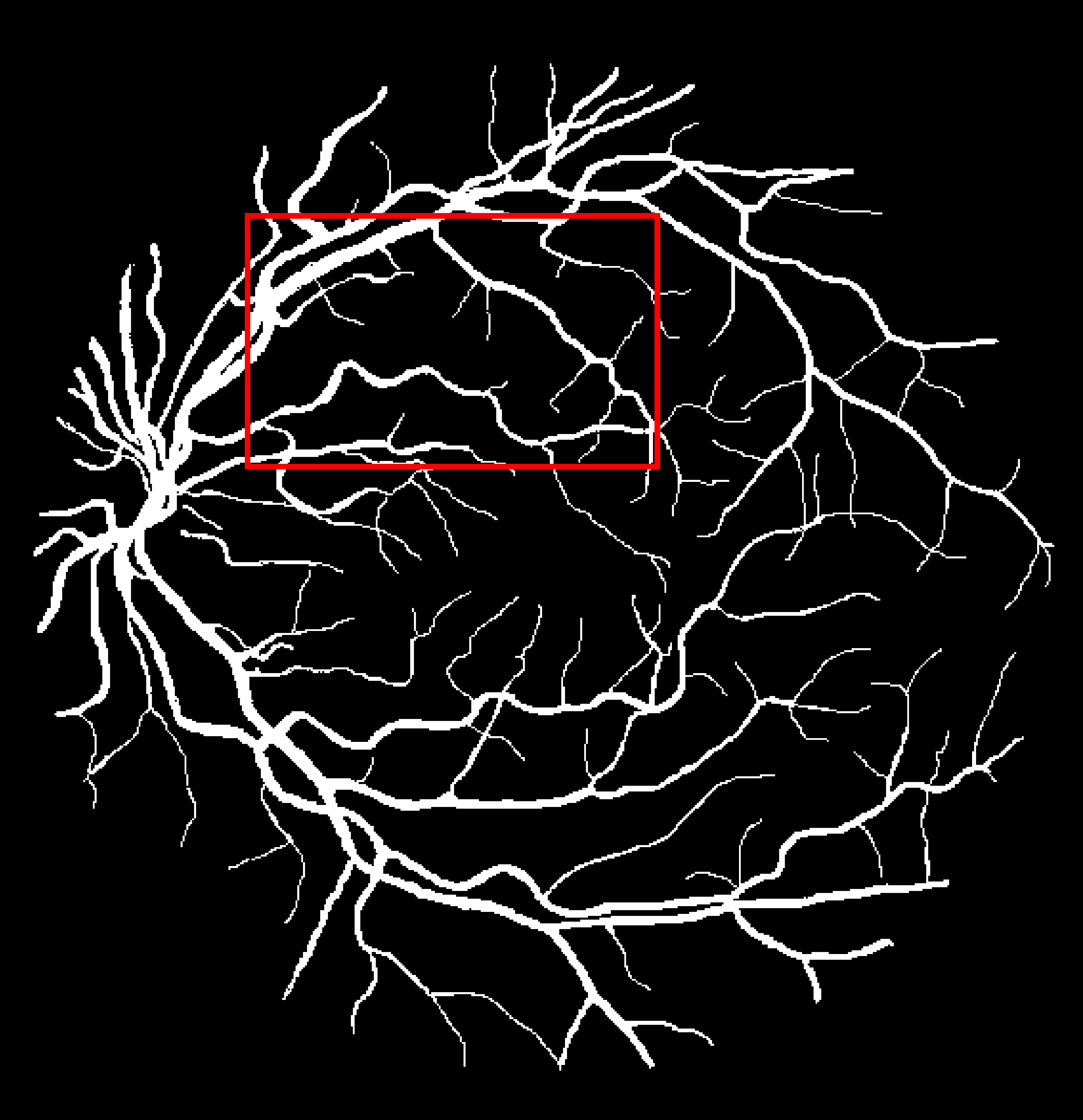} 
  \caption{} 
  \label{fig:experiment-gdt-01_test-drive}
  \end{subfigure}
  \quad
  \begin{subfigure}[b]{0.3\textwidth}
  \centering
  \includegraphics[height=.2\textheight]{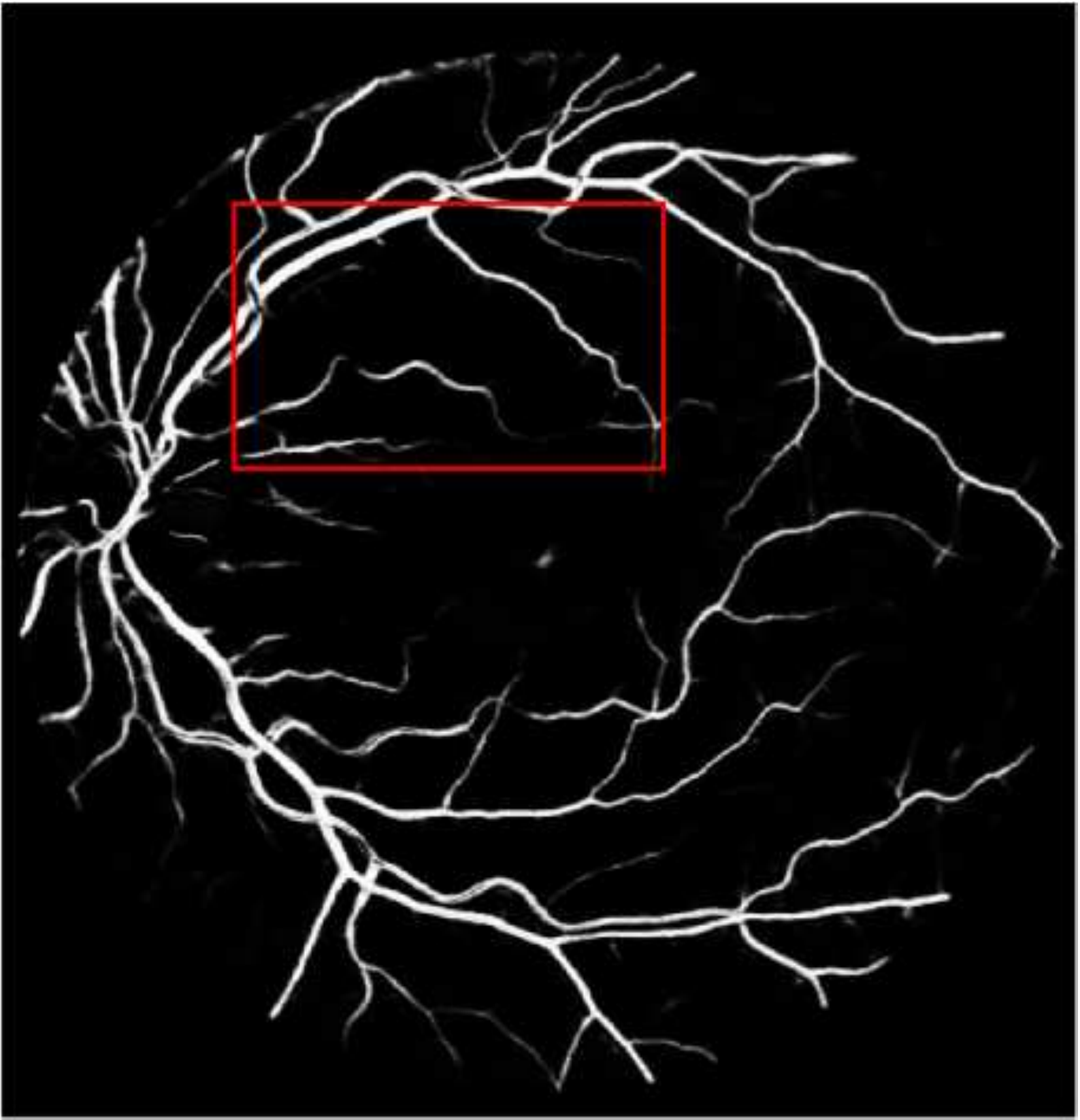} 
  \caption{} 
  \label{fig:experiment-thick-prob-drive}
  \end{subfigure}
  \quad
  \begin{subfigure}[b]{0.3\textwidth}
  \centering
  \includegraphics[height=.2\textheight]{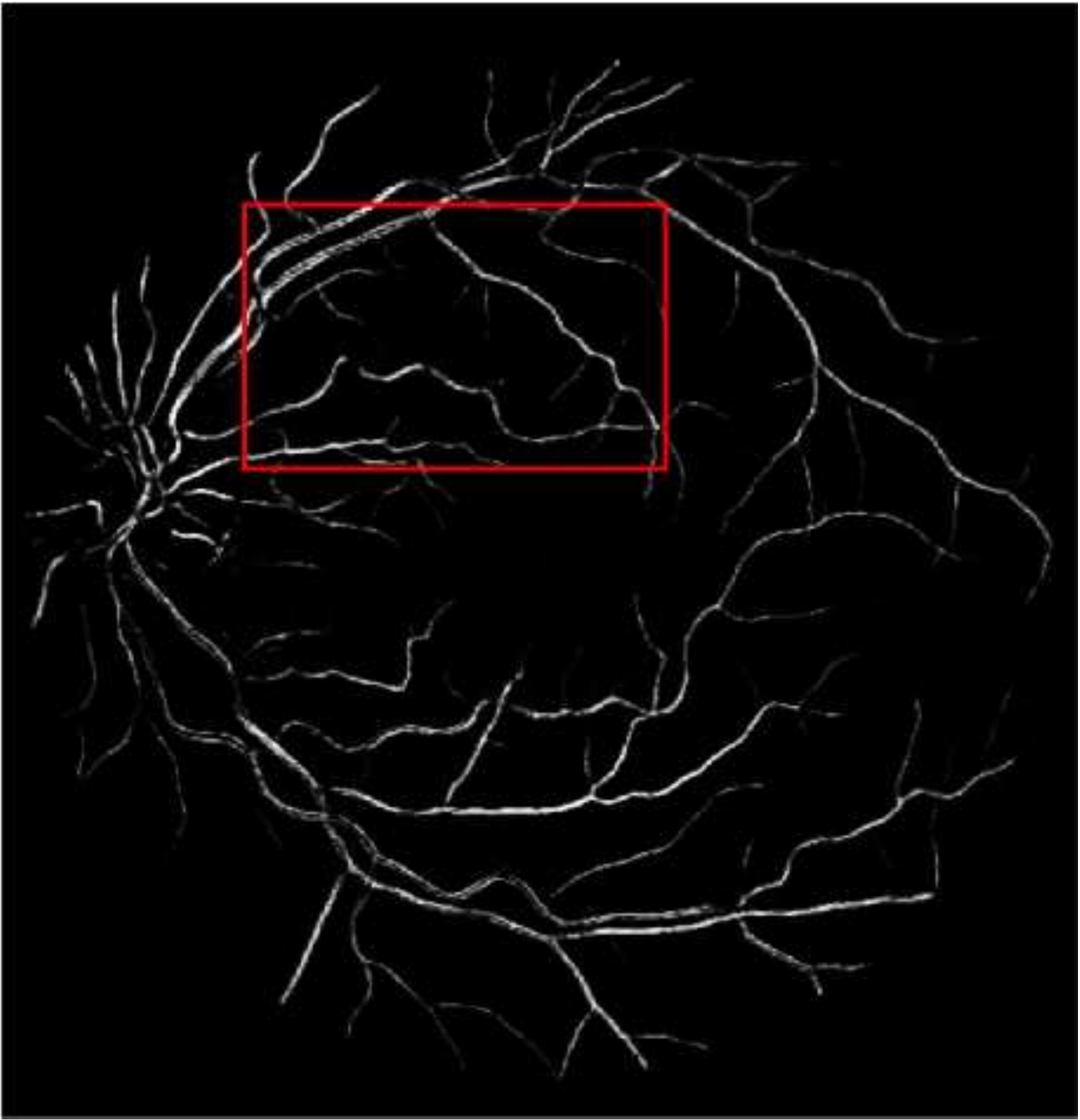} 
  \caption{} 
  \label{fig:experiment-thin-prob-drive}
  \end{subfigure}
  \\
  \begin{subfigure}[b]{0.45\textwidth}
  \centering
  \includegraphics[height=.2\textheight]{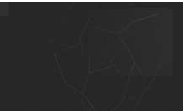}
  \caption{Detail of \figurename{\ref{fig:experiment-thin-sample-drive}}} 
  \label{fig:experiment-thin-magnified}
  \end{subfigure} 
  \quad
  \begin{subfigure}[b]{0.45\textwidth}
  \centering
  \includegraphics[height=.195\textheight]{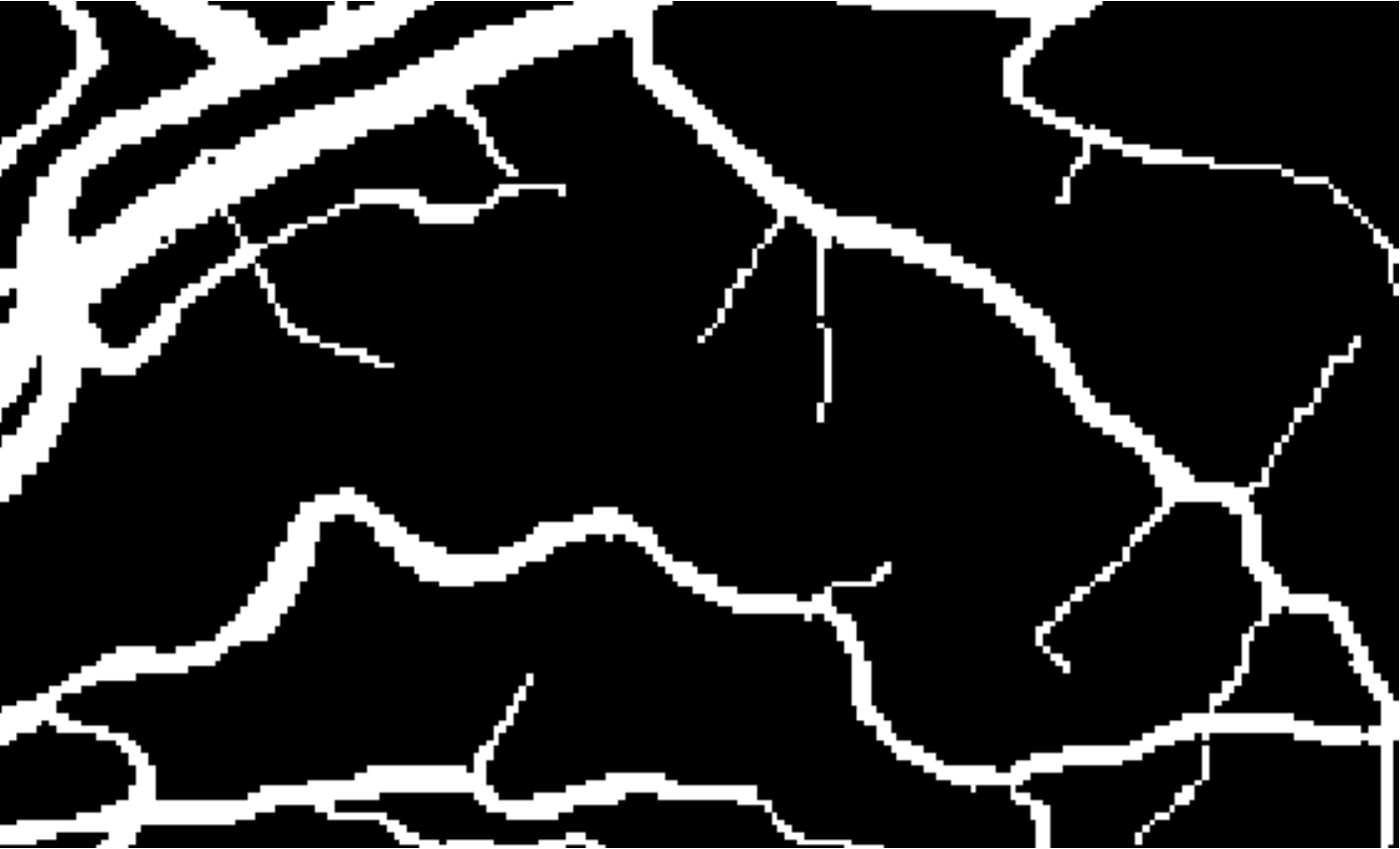}
  \caption{Detail of \figurename{\ref{fig:experiment-gdt-01_test-drive}} }
  \label{fig:experiment-gdt-details}
  \end{subfigure}
  \\
  \begin{subfigure}[b]{0.45\textwidth}
  \centering
  \includegraphics[height=.2\textheight]{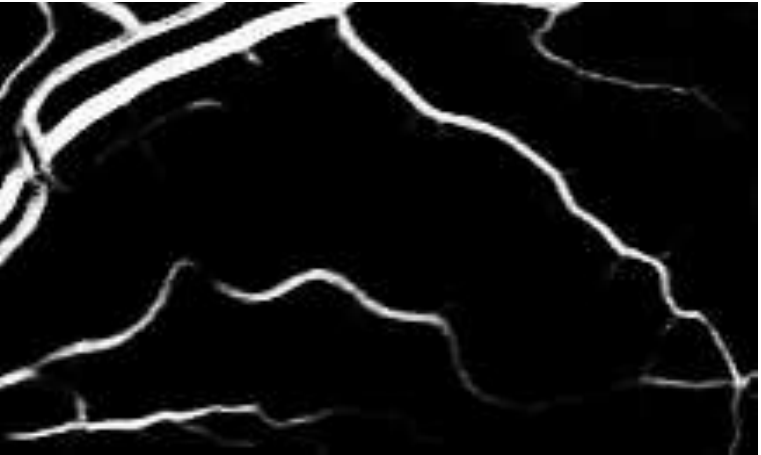}
  \caption{Detail of \figurename{\ref{fig:experiment-thick-prob-drive}}}
  \label{fig:experiment-continuity-magnified}
  \end{subfigure}
  \quad
  \begin{subfigure}[b]{0.45\textwidth}
  \centering
  \includegraphics[height=.2\textheight]{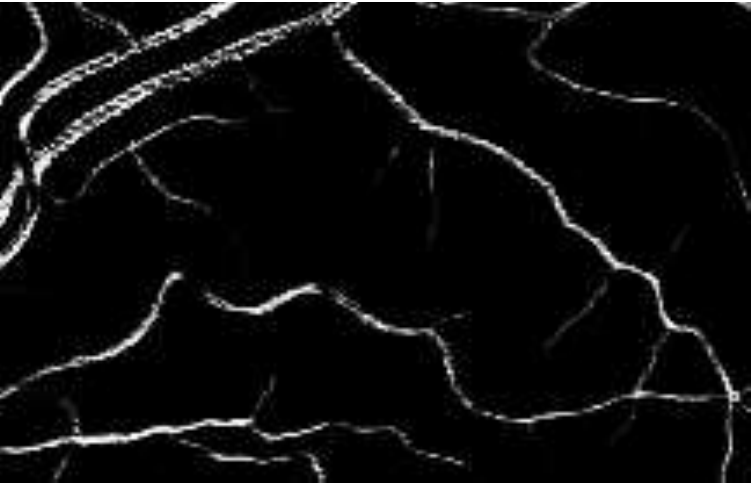}
  \caption{Detail of \figurename{\ref{fig:experiment-thin-prob-drive}}}
  \label{fig:experiment-discontinuity-magnified}
  \end{subfigure}
  \caption{Result comparison between different two datasets. (a) 
     Image from \textit{DRIVE} test data set (b) Sample from  \textit{dataset$\#1$}  
      (c) Sample from \textit{dataset$\#2$}
     (d) 	Ground truth  
     (e) Result by \textit{dataset$\#1$}
     (f) Result by \textit{dataset$\#2$}}
  \label{fig:experiment-sample-comparison}
\end{figure*}

In \figurename{\ref{fig:experiment-sample-comparison}}, we used
the two training datasets (\mbox{\textit{dataset$\#1$}}, \mbox{\textit{dataset$\#2$}})
to train the networks  respectively and 
tested the systems for same image. Intuitively, based on the 
constructions of the datasets, we thought \mbox{\textit{dataset$\#2$}} 
(\figurename{\ref{fig:experiment-thin-sample-drive}})
would yield
a more precise segmentation and the system would be more sensitive
for line-like patterns; however, since it only contains
thin lines the outputs would suffer from 
discontinuity.
The results follow our assumption
 (refer to \figurename{\ref{fig:experiment-sample-comparison}}).

\begin{table*}
\centering
\caption{Performance on \textit{DRIVE}  and \textit{STARE} database.}
\label{tab:performance-drive-stare}
\begin{adjustbox}{max width=\textwidth}
\begin{tabular}{cccccccccc}
\hline
\multirow{2}*{{Method}}  & \multicolumn{4}{c}{\textit{DRIVE}} &
 	& \multicolumn{4}{c}{ \textit{STARE}}		\\	
 	\cline{2-5}\cline{7-10}			& SN 	& SP 	& Acc & AUC  & & SN 	& SP 	& Acc & AUC \\
  \hline
  Marin \cite{marin2011a}    & .8139 & .9444 & .9308 &-  &  & .6776 & .9872 & .9689 &- \\
  Fraz \cite{fraz2012an}     &  .7406 & .9807 & .9480 & .9747 &
  &  .7548 & .9763 & .9534 & .9768 \\
  Zhao \cite{zhao2015automated}    & .742 & .982 & .954 & .862 &
  & .780 & .978 & .956 & .874 \\
  Roychowdhury  \cite{roychowdhury2015blood} & .7249 & .983 & .9519 & .962 &
  & .780 & .978 & .956 & .874 \\
  Azzopardi \cite{azzopardi2015trainable} & .7655 & .9704 & .9614 & .9442 &
  & .7716 & .9701 & .9563 & .9497 \\
   Zhang  \cite{zhang2016robust}    & .7473 & .9764 & .9474 & .9517 &
   & .7676 & .9764 & .9546 & .9614 \\
  Aslani  \cite{aslani2016a}       & .8302 & .9878 & 9688 & .9823 &
  & .8750 & .9874 & 9763 & .9925 \\
  Liskowski \cite{liskowski2016segmenting}  & .7763 & .9768 & .9495 & .9720 &
  & .7867 & .9754 & .9566 & .9785 \\
   Farokhian  \cite{Farokhian2017} & .6933 & .9777 & .9392 & .9530 & & - & -& - & -\\
  Pandey \cite{Pandey2017162} & .8106 & .9761 & .9623 &.9650 & & .8319 & .9623 & .9444 & .9547\\
  {\bf Our result} & .7426 & .9735 & .9453 & .9516 & & .7295 & .9696 & .9449 &.9557 \\

\hline
\end{tabular}
\end{adjustbox}
\end{table*}

\begin{table}
\centering
\caption{Performance on \textit{DRIVE} DATASET.}
\label{tab:performance-on-drive}
\begin{tabular}{ccccccc}
\hline
Image & AUC & Acc & & Image & AUC & Acc   \\
\hline
01 & .9425 & .9289 & & 11 & .8871 & .9172 \\
02 & .9054 & .9043 & & 12 & .9109 & .9163 \\
03 & .9332 & .9194 & & 13 & .8963 & .9050 \\
04 & .8975 & .8219 & & 14 & .9170 & .9278 \\
05 & .9170 & .9181 & & 15 & .9347 & .9412 \\
06 & .9061 & .9130 & & 16 & .8934 & .9123 \\
07 & .8927 & .9132 & & 17 & .8791 & .9035 \\
08 & .8962 & .9097 & & 18 & .9026 & .9181 \\
09 & .8767 & .9019 & & 19 & .9516 & .9452 \\
10 & .9268 & .9316 & & 20 & .9382 & .9331 \\  
\hline
\end{tabular}
\end{table}

It is also worth noting that, in this work, we do not focus on tricks 
in training deep neural networks, 
just a naive structure and a short iterating (within $200,000$ iterations, $2$ samples for each)
 with default settings,
however, the system has achieved competitive performance
 (see \tablename{\ref{tab:performance-drive-stare}}).
 Also, we give a comprehensive view of the nets' performance on \textit{DRIVE}
 dataset (\tablename{\ref{tab:performance-on-drive}}).

\section{Conclusions}

In this work, we focused on using prior knowledge to leverage
the great power of deep neural networks in feature learning 
for retinal blood vessel segmentation.
The study is based on the fact that although the deep neural networks
has been proved to be a powerful tool, the high cost of labeling 
hampers its applications in vast areas.
We argue that, with strong domain-specific prior knowledge, we
are able to drive the deep neural networks to an alternative direction 
where a desirable performance is available.

Based on this assumption, we propose a novel approach to 
supervising deep neural networks for blood
vessel  segmentation in fundus images which is free from manual labeling.
 In our approach, we pour our prior knowledge
into the dataset to generate artificial training samples and labels which are
used to train the deep neural networks. To illustrate our
approach, we constructed the training samples with line segments; and then,
 we built a naive deep neural networks model and trained it on the artificial images 
 in a straightforward way.
 
 The results  have shown that 
 our approach achieved competitive performance on benchmark datasets,
 and provide considerable evidence for our hypothesis.
 As deep learning is still an ongoing study which is, currently,
  mainly based on intuition and engineering, we cannot strictly prove 
  this hypothesis with this one application; but this work implies a potential way
  to leverage deep neural networks without manual labeling.

\bibliography{ref}
\end{document}